\documentclass[letterpaper,10pt,conference]{IEEEtran}

\usepackage{ifthen}
\newboolean{arxiv}
\setboolean{arxiv}{true}   

\IEEEoverridecommandlockouts

\usepackage[table]{xcolor}
\definecolor{maroon}{cmyk}{0,0.87,0.68,0.32}

\usepackage{fontawesome}
\usepackage{pifont}

\usepackage{listings}

\usepackage{rotating}

\usepackage{tikz}
\usetikzlibrary{spy,calc}

\usepackage{tabularx}
\usepackage{caption}

\usepackage{graphicx}

\usepackage{amssymb}
\usepackage{amsmath}

\usepackage{color}
\usepackage{colortbl}

\usepackage[binary-units]{siunitx}
\DeclareSIUnit\mph{mph}

\usepackage[hidelinks]{hyperref}
\newcommand\rurl[1]{%
\texttt{\href{http://#1}{\nolinkurl{#1}}}%
}

\usepackage{cleveref}
\crefname{table}{Table}{Tables}
\crefname{figure}{Figure}{Figures}
\crefname{section}{Section}{Sections}
\crefname{equation}{Equation}{Equations}

\usepackage{crossreftools}

\usepackage{todonotes}

\begin{document}

\bstctlcite{IEEEexample:BSTcontrol}

\title{
{
\Large \bf
\textit{Open-RadVLAD}: Fast and Robust Radar Place Recognition
}
}
\author{Matthew Gadd and Paul Newman\\
Mobile Robotics Group, University of Oxford\\
{\faEnvelope} \texttt{mattgadd@robots.ox.ac.uk}
{\faGithub}~\rurl{github.com/mttgdd/open-radvlad}
}
\maketitle

\begin{abstract}
Radar place recognition often involves encoding a live scan as a vector and matching this vector to a database in order to recognise that the vehicle is in a location that it has visited before.
Radar is inherently robust to lighting or weather conditions, but place recognition with this sensor is still affected by: (1) viewpoint variation, i.e. translation and rotation, (2) sensor artefacts or ``noises''.
For \SI{360}{\degree} scanning radar, rotation is readily dealt with by in some way aggregating across azimuths.
Also, we argue in this work that it is more critical to deal with the richness and informativeness of representation than it is to deal with translational invariance -- particularly in urban driving where vehicles predominantly follow the same lane when repeating a route. 
In our method, for computational efficiency, we use only the polar representation.
For partial translation invariance, we use only a one-dimensional Fourier Transform along radial returns.
As the original radar signal is in the form of received power in discretised range bins, we also show experimentally that taking a radial Fourier Transform in this way and matching based on spatial frequencies present in the power signal leads to better performance -- leading to a \SIrange{7}{8}{\percent} improvement in localisation success (\cref{sec:experiments,tab:average_results}).
We also achieve rotational invariance and a very discriminative descriptor space by building a vector of locally aggregated descriptors (VLAD).
Our method is more comprehensively tested than all prior radar place recognition work -- over an exhaustive combination of all $870$ pairs of trajectories from $30$ \textit{Oxford Radar RobotCar Dataset} sequences (each $\approx$\SI{10}{\kilo\metre}), with a frequency-modulated continuous-wave (FMCW) radar.
Code and detailed results are provided at
\rurl{github.com/mttgdd/open-radvlad}, as an open implementation and benchmark for future work in this area. 
We achieve a mean of \SI{89.35}{\percent} and median of \SI{91.52}{\percent} in \texttt{Recall@1}, outstripping the mean of \SI{68.56}{\percent} and median of \SI{69.55}{\percent} for the only other open implementation, \ref{V2}, and at a fraction of its computational cost (relying on fewer integral transforms e.g. Radon, Fourier, and inverse Fourier).
\end{abstract}

\section{Introduction}
\label{sec:intro}

For autonomous vehicles to drive in a safe way even in the face of challenging illumination or weather conditions very robust sensing is required.
Thus, the interest in scanning frequency-modulated continuous-wave radar for place recognition and localisation.
However, radar place recognition is not an especially mature research area -- there are few open-source implementations and no exhaustive evaluations.
Therefore, in this work, our principal contributions are:
\begin{enumerate}
\item A novel \textbf{simplification} of recent work using integral transform techniques -- using only a single forward Fourier Transform, with better place recognition \textbf{performance} and with much faster computation,
\item An \textbf{exhaustive evaluation} over the \textit{Oxford Radar RobotCar Dataset} -- the largest such evaluation to date, and
\item An \textbf{open implementation}, to serve as a platform for future research.
\end{enumerate}

\begin{figure}
\centering
\includegraphics[width=\columnwidth]{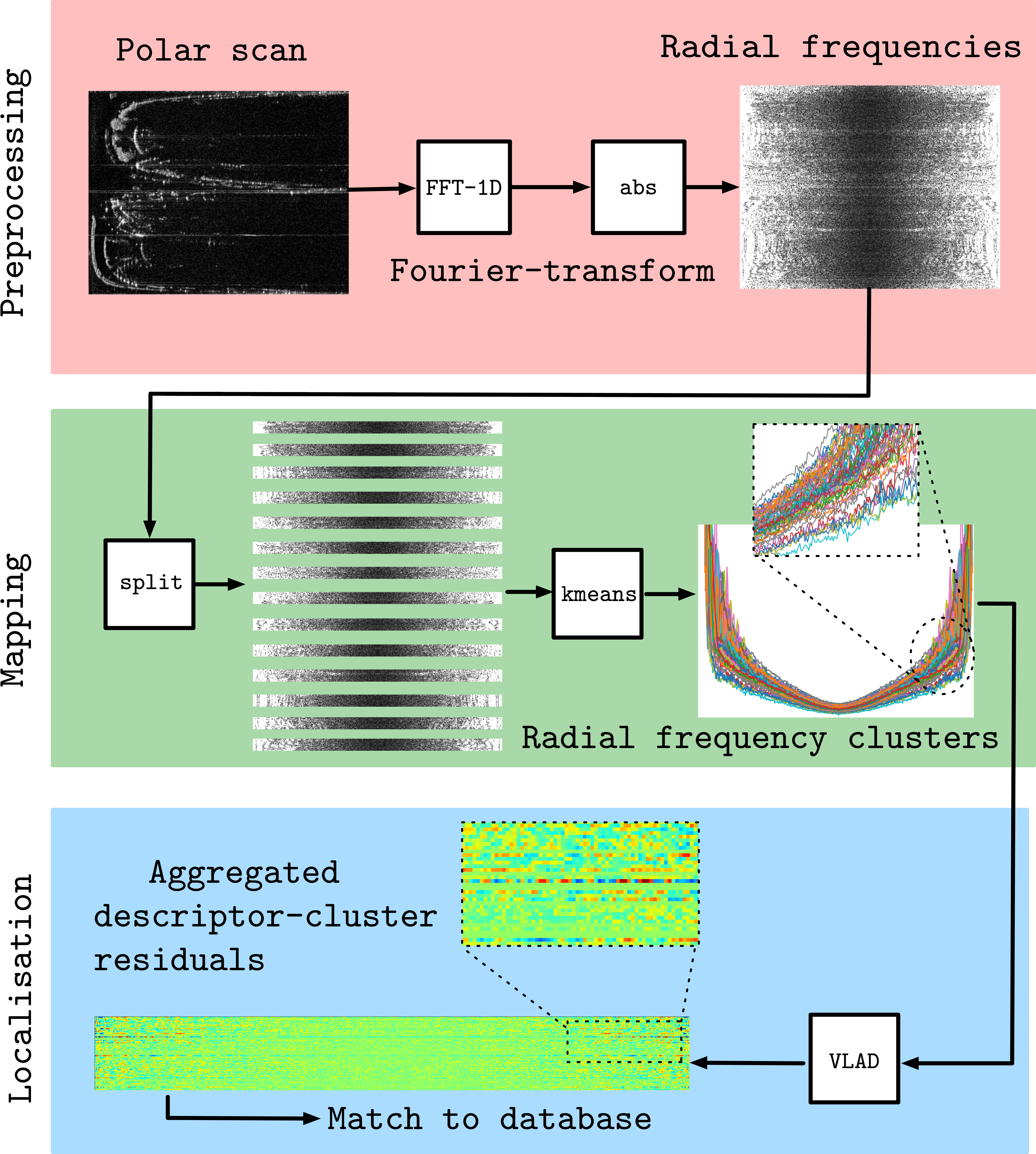}
\caption{
\textit{Open-RadVLAD system overview}.
For a degree of robustness, we take the 1D Fourier Transform of the radial returns.
Clustering these radial frequency responses is not specific to azimuth order and so gives us rotational invariance.
Therefore, we apply a ``vector of locally aggregated descriptors'' (VLAD) as an informative scan descriptor with the residuals of query radial frequency responses to those cluster centres.
}
\label{fig:system}
\end{figure}

\cref{fig:system} gives an overview of our system.
For background, place recognition can be performed by matching a vector representation of radar scans to vectors from previously visited places (indicated by \texttt{Match to database}).
This paper is about \textit{engineering good representations} -- with incorrect matches ``far away'' and correct matches ``close by'' in that vector space.

\section{Related Work}
\label{sec:related_work}

Several recent radar datasets use the same radar class -- from the same manufacturer\footnote{\textit{Navtech Radar}: \rurl{navtechradar.com}} -- as used in our work, including the \textit{Oxford Radar RobotCar Dataset}~\cite{barnes2020oxford} by Barnes \textit{et al}, \textit{MulRan}~\cite{kim2020mulran} by Kim \textit{et al}, \textit{RADIATE}~\cite{sheeny2021radiate} by Sheeny \textit{et al}.
The \textit{Oxford Radar RobotCar Dataset}~\cite{barnes2020oxford} is purely urban and features many repeat traversals, useful in investigating place recognition.

Radar place recognition has been explored 
in~\cite{suaftescu2020kidnapped,barnes2020under,gadd2020look,wang2021radarloc,komorowski2021large,gadd2021contrastive,hong2022radarslam,adolfsson2023tbv,jang2023iros,yuan2023iros}.
This has included simultaneous localisation and mapping techniques~\cite{hong2022radarslam,adolfsson2023tbv} methods for learning to vectorise radar scans with neural networks 
in~\cite{suaftescu2020kidnapped,barnes2020under,gadd2021contrastive,wang2021radarloc,komorowski2021large,yuan2023iros}
and also non-learned 
methods~\cite{kim2020mulran,gadd2020look,jang2023iros}. 

The most relevant to our work is \textit{RaPlace} by Jang \textit{et al}~\cite{jang2023iros} as originally applied to LiDAR place recognition in \textit{RING} by Lu \textit{et al} in~\cite{lu2022one}.
In these works, a Radon Transform~\cite{beylkin1987discrete} on the Cartesian scan gives a Sinogram, and a Fourier Transform of this Sinogram gives a descriptor which is totally insensitive to translation of the Cartesian scan, and shifts circularly under rotation of the Cartesian scan.
Then, circular cross-correlation is used to measure the similarity between descriptors in a rotationally-invariant way.
This circular cross-correlation itself involves two Fourier Transforms and an inverse Fourier Transform.

Our experimental results in~\cref{sec:results} indicate that a much greater impact on localisation performance is available by focusing on the richness of the vector representation rather than translation invariance.
We do, however, as detailed further in~\cref{sec:method} below, use the Fourier Transform directly on polar radar returns (rather than a Sinogram obtained from the Cartesian scan which itself is obtained from the polar scan).

\section{Method}
\label{sec:method}

As shown in the top left of~\cref{fig:system} by \texttt{Polar~scan}, we start with range-bearing arrays $f(r,\theta)\in\mathbb{R}^{W\times{}H}$ where $W$ is the number of range bins $r$ (horizontal axis) and $H$ is the number of discrete azimuths $\theta$ (vertical axis).

Then, we then take the 1D Discrete Fourier Transform along the radial dimension
\begin{equation}\label{eq:fft}
\hat{f}(\rho,\theta)=\sum_{r=0}^{W-1}f(r,\theta)e^{-\frac{i2\pi}{W}\rho{}r}
\end{equation}
giving radial frequency responses along each heading.
Although as mentioned in~\cref{sec:intro} these radial returns are themselves powers, the use of the Fourier Transform with the \textit{Navtech} class of radar sensor (\cref{sec:experiments}) has been shown to be effective in e.g.~\cite{weston2022fast} -- and similarly for the related Fourier Mellin Transform (FMT) in~\cite{park2020pharao} -- and is proved in~\cref{sec:experiments} to give an important boost to place recognition performance.
We specifically use the magnitude of this complex number $\hat{f}_\mathrm{abs}(\rho,\theta)=|\hat{f}(\rho,\theta)|$.
These steps are shown as \texttt{FFT-1D} and \texttt{abs} in~\cref{fig:system}.
Each of these radial frequency response magnitudes is a $W$-dimensional vector $\hat{f}_\mathrm{abs}(\rho,\theta_j)$ specific to a heading $\theta_j$, shown after the \texttt{split} in~\cref{fig:system}.

Now, with more detail below, localisation consists of matching a sequence of query scans to a sequence of reference scans -- the latter being referred to as a ``map''.
We now take all radial frequency responses from all scans in a map trajectory\footnote{
e.g. if there were $1000$ scans in a map, each with $400$ azimuths, we perform clustering on $400\times1000=400000$ vectors
} (or ``experience'') and find cluster centres
$
C=\{\mathbf{c}_1, \mathbf{c}_2,\ldots,\mathbf{c}_k\}
$
where each $\mathbf{c}_i\in\mathbb{R}^W$ is in the space formed by all $\hat{f}_\mathrm{abs}$ from all azimuths from all scans in the map. 
Examples of these radial frequency clusters are shown in~\cref{fig:system} (middle, green).

Now, as in~\cref{eq:nn} below, to find matches between the query experience and map experience, we convert polar map scans as well as polar query scans into some vector representation between which we can compute distances.
For this, we use the vector of locally aggregated descriptors~\cite{arandjelovic2013all}. Here, each contiguous section of the ``vector of locally aggregated descriptors'' (VLAD) descriptor $\mathbf{v}_{i}$ is computed as the sum of residuals from every radial frequency response which has its nearest cluster (written $\mathrm{nn}$ in~\cref{eq:vlad}) the centre corresponding to that section, as per
\begin{equation}\label{eq:vlad}
\mathbf{v}_{i}=\sum_{\mathbf{x}_j|\mathrm{nn}(\mathbf{x}_j)=\mathbf{c}_i}(\mathbf{x}_j-\mathbf{c}_i),\quad
\mathbf{x}_j=\hat{f}_{\mathrm{abs}}(\rho,\theta_j)
\end{equation}
and the full VLAD descriptor $\mathbf{v}$ is found by concatenating $\mathbf{v}_{i}$ for all $\mathbf{c}_i$.
An example is shown in~\cref{fig:system} (bottom, blue).

With map VLAD descriptors $\mathbf{v}^j$ at places indexed by $j$ and query VLAD descriptor $\mathbf{w}$, localisation can then be stated as
\begin{equation}\label{eq:nn}
j^*=\operatorname{argmin}_j||\mathbf{w}-\mathbf{v}^j||_2^2
\end{equation}
where we decide that we are in place $j^*$ given that the descriptor for that place is closest to our query descriptor (i.e. ``place recognition'').

Note that we may also accept the \texttt{N} nearest neighbours as a range of $j^*$ values for putative localisation candidates, as we do in our experiments in~\cref{sec:experiments,sec:results}.
With more than $1$ candidate, it is often expected as in e.g.~\cite{kim2020mulran} that some downstream pose refinement module disambiguates them and chooses the correct result, with hierarchical interaction between the ``place recognition'' and ``pose/orientation refinement'' modules (in reality: two distinct localisation systems).
This, however, is not the topic of this work -- we are purely interested in constructing a good representation of polar scans with vectors/descriptors that represent ``places'' well, and further pose refinement would benefit our approach equally.

\begin{figure*}[t]
\centering
\includegraphics[width=0.49\textwidth]{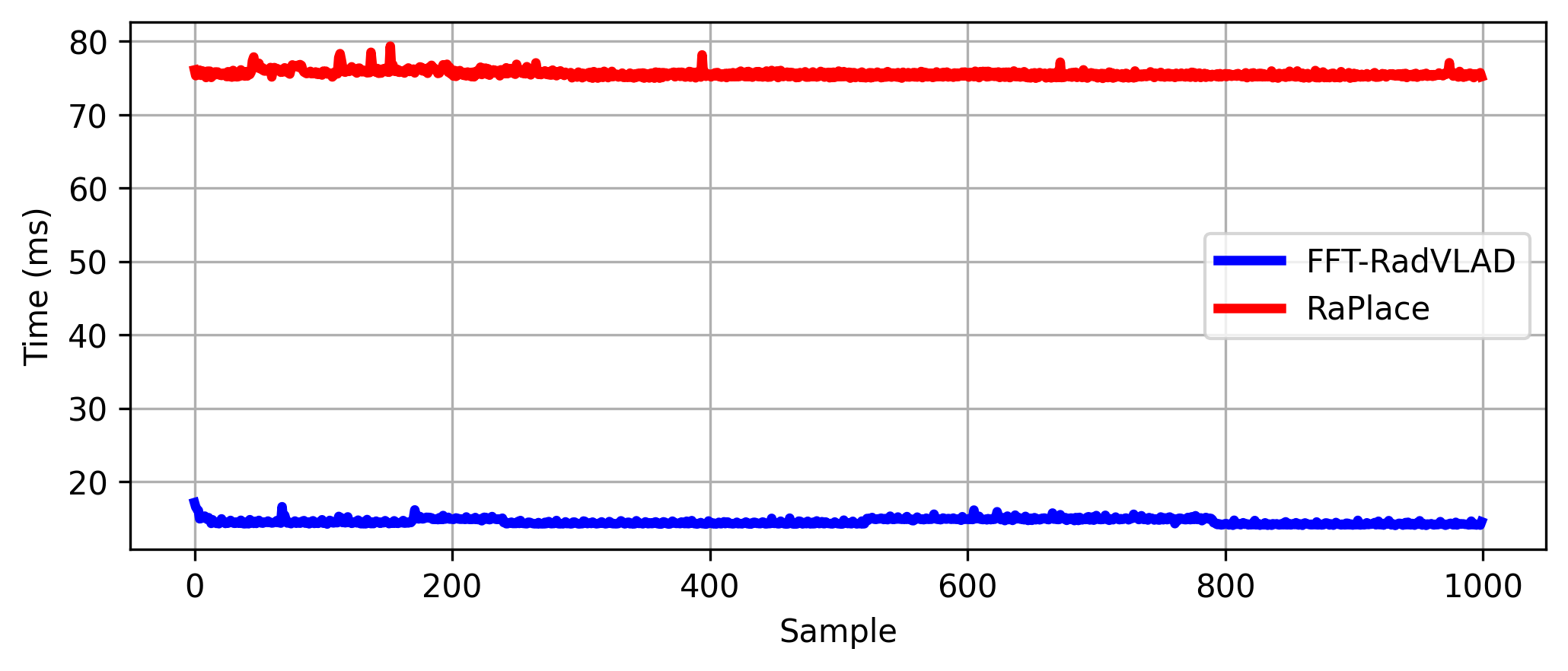}
\includegraphics[width=0.49\textwidth]{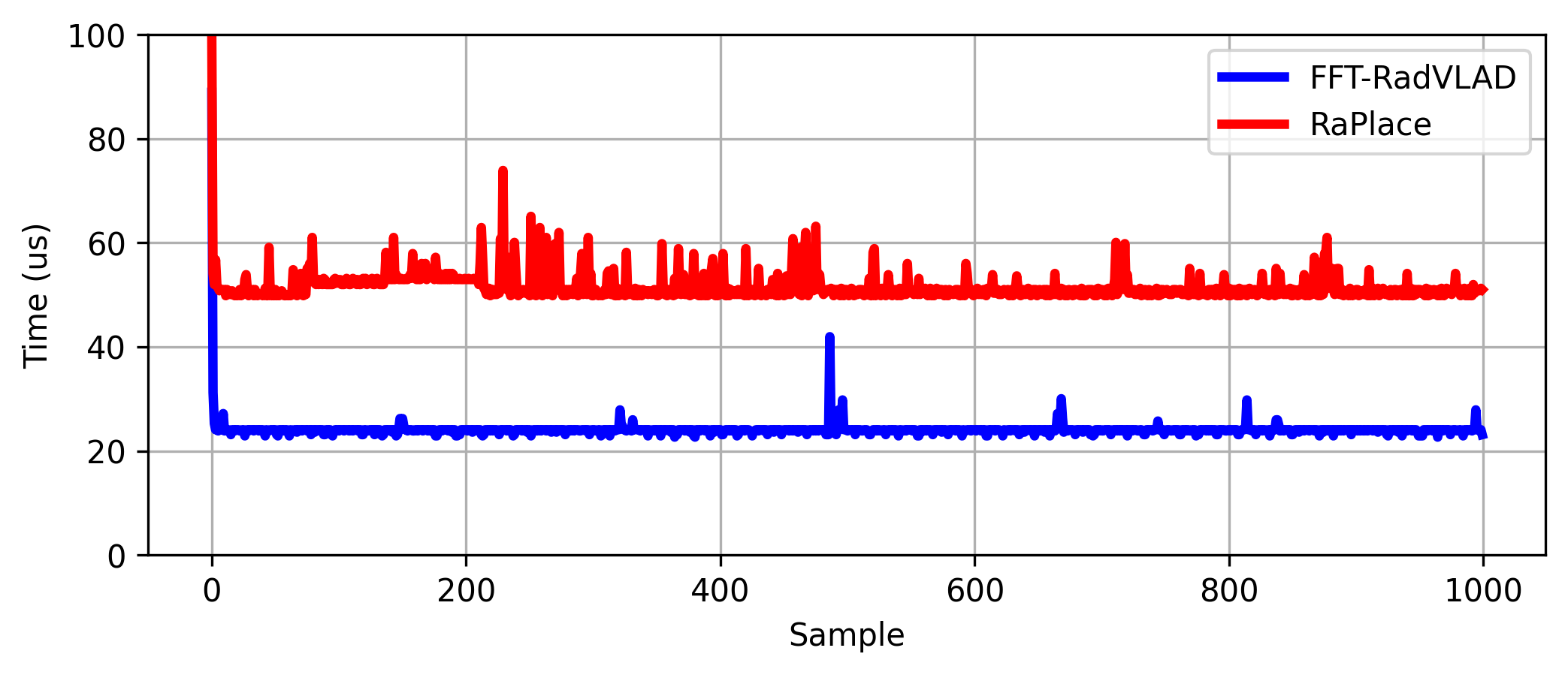}
\caption{
\textit{Top}: $1000$ samples of time taken to \textbf{build representations} for \ref{V4}, our method (blue), as compared to \ref{V2} from~\cite{jang2023iros} (orange).
\textit{Bottom}: $1000$ samples of time taken to \textbf{compute representation distance}.
These samples were collected by running processes on an \textit{Apple M2 Pro} with $12$ cores and with \SI{16}{\giga\byte} \textit{LPDDR5} memory.
}
\label{fig:timing}
\end{figure*}

The motivation for our approach as outlined above is
\begin{enumerate}
\item \textbf{Polar scans only}: We stay in the natural signal form for scanning radar and do not map to Cartesian form, in order to keep computation to a minimum, avoid interpolation artefacts, and minimise the number of settings and hyperparameters.
\item \textbf{Spatial frequencies in power returns}: In implementing~\cref{eq:fft} we use the Fast Fourier Transform from \texttt{numpy}\footnote{\rurl{numpy.org/doc/stable/reference/routines.fft.html}}.
We show in~\cref{sec:experiments} that this leads to a marked improvement in localisation success than if we were to match radial power returns themselves.
\item \textbf{Discriminativeness}: As~\cref{eq:vlad} is based on residuals to cluster centres, the VLAD descriptor effectively captures information about the distribution of radial frequency responses \textit{relative to each other} (not just their values), and this higher-order information is important for uniquely representing inputs.
\item \textbf{Rotation invariance}: In~\cref{eq:vlad}, the order in which radial frequency responses at particular headings are mapped to a nearest cluster and then aggregated is irrelevant.
\end{enumerate}

Note that in comparison to \textit{RaPlace}~\cite{jang2023iros}, even though shifts along one radial return do not feature in the magnitude of our Fourier Transform in~\cref{eq:fft}\footnote{
For $g(r,\theta_j)=f(r+\delta{}r,\theta_j)$ we have $\hat{g}(\rho,\theta_j)=e^{+j\rho\delta{}r}\hat{f}(\rho,\theta_j)$ with some $\delta{}r$ along a particular $\theta_j$
}, we do not have total shift invariance using a Radon and Fourier transform combination.
Also, we rely on very high-dimensional descriptors (with length equal to the number of range bins $\times$ the number of cluster centres).
However, there are two important points to consider in comparing our method to~\cite{jang2023iros}.
Firstly, the informativeness of our longer descriptor leads to \textbf{much bigger performance benefits} (see~\cref{fig:recall_plot}, with experimental verification in in~\cref{sec:experiments,sec:results} below).
Secondly, and critically, despite the length of our descriptor, our method is \textbf{much more computationally efficient} than~\cite{jang2023iros} (see~\cref{fig:timing} \textit{Right}, explained further in~\cref{sec:results} below), as our vectors are compared with simple Euclidean distances rather than circular cross-correlation which itself requires two Fourier transforms and an inverse Fourier Transform.
We also build representations more quickly than~\cite{jang2023iros} (\cref{fig:timing} \textit{Left}), as we do not use polar-to-Cartesian mapping or the Radon Transform prior to the Fourier Transform.

\section{Experiments}
\label{sec:experiments}


\subsection{Dataset}
\label{sec:dataset}

We evaluate our method in training and testing across urban data collected in the \textit{Oxford Radar RobotCar Dataset}~\cite{barnes2020oxford}.
The dataset was collected using an autonomous-capable \textit{Nissan LEAF}.
This follows the original \textit{Oxford RobotCar Dataset}~\cite{maddern20171} route and over $32$ traversals in different traffic, weather and lighting conditions, totalling $\approx$\SI{280}{\kilo\metre} of driving.
This features a \textit{CTS350-X Navtech} frequency-modulated continuous-wave (FMCW), scanning radar with operating frequency range \SIrange{76}{77}{\giga\hertz} and scan rate is \SI{4}{\hertz}.
The \textit{CTS350-X Navtech} rotates about its vertical axis while transmitting and receiving frequency-modulated
radio waves.
Scans have \SI{3768}{} range bins at a resolution of \SI{4.38}{\centi\meter}, with total range \SI{165}{\metre}.
There are \num{400} azimuths (resolution \SI{0.9}{\degree}).

In contrast to all prior radar place recognition work, we use \textit{all} trajectories from this dataset, with the exception of only \texttt{-partial} completions of the central Oxford route.
The $30$ ``experiences'' used are listed in~\cref{tab:average_results}.
\ifthenelse{\boolean{arxiv}}{Full results are available in~\cref{tab:full_results}.}{}
There are a total of $30^2-30=870$ pairs of unique experiences.
For each pair, we use one experience as the \textit{reference} trajectory or map to localise to and the other experience for \textit{query} frames.
This is the most comprehensive evaluation of place recognition over this dataset to date.

A foray typically consists of \SIrange{8000}{9000}{} scans.
We \textit{downsample} the framerate, taking every $10^{\text{th}}$ scan.
Thus, each experience features $\approx$\SIrange{800}{900}{} places.
Note that this corresponds to about $\frac{10}{4}=\SI{2.5}{\second}$ during which time at \SI{20}{\mph} or \SI{8.9408}{\metre\per\second} the vehicle travels about $8.9408\times2.5=\SI{22.352}{\metre\per\second}$ -- the typical distance between consecutive places.

\subsection{Performance metrics}

\begin{figure}[!h]
\centering
\begin{tikzpicture}[spy using outlines={circle,red,magnification=3,size=2cm, connect spies}]
\node [inner sep=0pt] (matgt) at (0,0){\includegraphics[width=0.4\columnwidth]{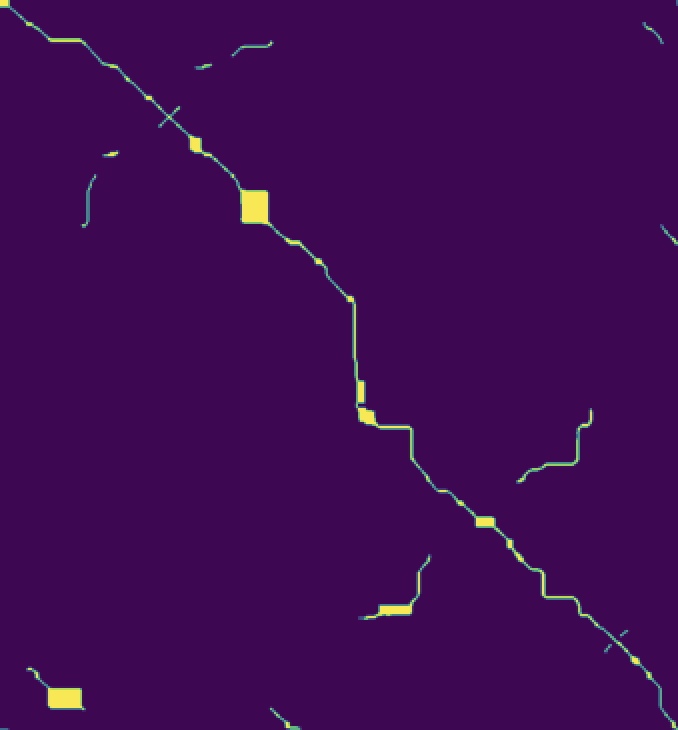}};
\node [inner sep=0pt] (matv4) at (4,0){\includegraphics[width=0.4\columnwidth]{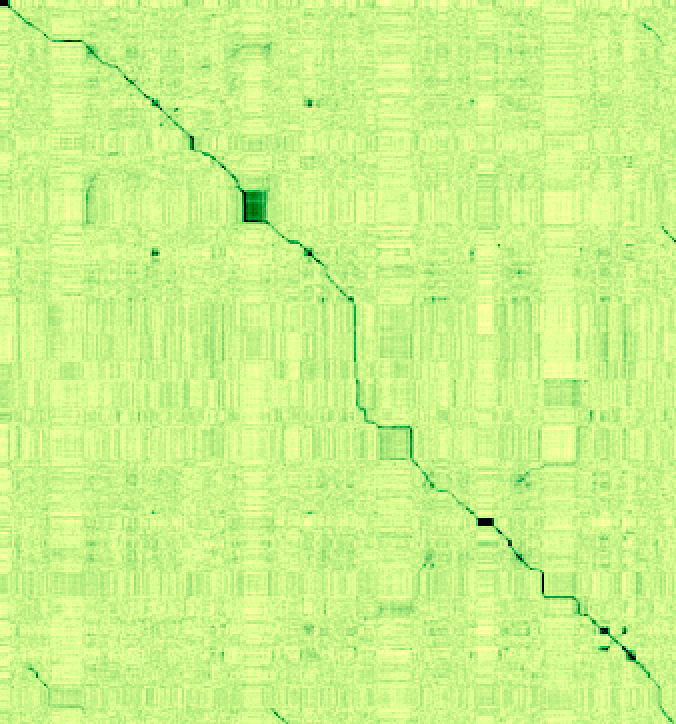}};
\spy on (2.75,0.85) in node [right] at (5.25,1);
\spy on (2.5,-1.6) in node [right] at (5.25,-1.1);
\end{tikzpicture}
\caption{
\textit{Left}: Ground truth GPS difference matrix for the $\approx$$800\times800$ possible place correspondences between sequences \texttt{2019-01-18-15-20-12} and \texttt{2019-01-15-12-01-32}.
\textit{Right}: Corresponding embedding distance matrix for \ref{V4}.
Red insets show revisits in the opposite lane/direction (top) and in the same lane/direction (bottom).
}
\label{fig:mats}
\end{figure}

To assess the place recognition performance, we use \texttt{Recall@N (R@N)}, which measures the percentage of query frames which have a nearest neighbour in the map which is actually close in physical space.
For this we use a ``difference matrix'' in~\cref{fig:mats} \textit{Right} with the embedding distances (Euclidean) between live and map features.
Localisation is successful if the nearest neighbour for a query embedding (rows) is a reference embedding (column) which is in truth is close to the query location in physical space.
For this ground truth, in~\cref{fig:mats} \textit{Right}, we use the GPS/INS provided in the \textit{Oxford Radar RobotCar Dataset} (a NovAtel SPAN-CPT ALIGN inertial and GPS navigation system), and consider a match to be good if it is within \SI{25}{\metre}  -- a very commonly accepted threshold, e.g. as is used in~\cite{arandjelovic2016netvlad,berton2022rethinking}.

\subsection{Baselines \& ablation}
\label{sec:baselines}

The methods for comparison are:
\begin{enumerate}
\item[\ding{172}] {\crtcrossreflabel{\texttt{RingKey}}[V1]}: from~\cite{kim2018scan,kim2020mulran}\footnote{
Open implementation at \rurl{github.com/irapkaist/scancontext/tree/master/fast_evaluator_radar}
}.
This method consists of turning the polar array into a vector by simply finding the average of returns at distinct ranges around the vehicle -- e.g. collapsing \cref{fig:polar} into a single row.
Note that we do not use the additional orientation refinement from~\cite{kim2018scan}, as distances between row-reduced vectors and orientation scores are separate, i.e. these form a hierarchy of localisers as mentioned in~\cref{sec:method} and we are interested in comparing the engineering of vector representations. 
\item[\ding{173}] {\crtcrossreflabel{\texttt{RaPlace}}[V2]}: from~\cite{lu2022one}\footnote{
Open implementation at \rurl{github.com/hyesu-jang/RaPlace/blob/main/PYTHON/RaPlace.py}
} as described in~\cref{sec:method}.
Through maximising circular cross-correlation, this method actually does recover orientation.
However, in contrast to \ding{173}, this orientation score is what is used for place recognition/similarity measurement.
\item[\ding{174}] {\crtcrossreflabel{\texttt{RadVLAD}}[V3]}: from~\cref{sec:method}.
This involves VLAD descriptors but \textit{no Fourier Transform} (i.e. we cluster, etc the raw radar signals).
\item[\ding{175}] {\crtcrossreflabel{\texttt{FFT-RadVLAD}}[V4]}: from~\cref{sec:method}.
This is as per \ding{174} but in this case we use the radial frequency responses rather than raw radar signals.
\end{enumerate}

Here, \ding{172} and \ding{173} are competitors, while \ding{174} and \ding{175} serve as an ablation study for the two critical components of our work.

Note that the evaluation of \ref{V2} in~\cite{jang2023iros} on the \textit{Oxford Radar RobotCar Dataset} was restricted to experiences from \texttt{2019-01-16} and \texttt{2019-01-18}, while our evaluation covers the entire month of data.
Also in contrast to~\cite{jang2023iros}, we do not look for loop closures \textit{within an experience} -- which may be scarce -- instead focusing on the inter-experience setup, where loop closures are abundant (every frame in the query sequence has matches in the reference sequence).

\subsection{Settings \& hyperparameters}

\begin{figure}
\centering
\begin{tikzpicture}[spy using outlines={circle,red,magnification=5,size=1.25cm, connect spies}]
\node {\includegraphics[width=0.5\columnwidth]{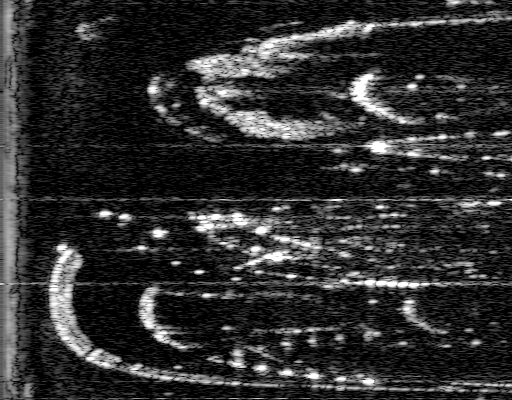}};
\spy on (-2.1,0) in node [left] at (-3.6,1.25);
\end{tikzpicture}
\caption{
\textit{Example polar scan}, with maximum range limited to $1024$ pixels and then resized to $400\times512$.
The zoomed in inset (red) shows \textbf{returns from near the vehicle} (first few range columns).
}
\label{fig:polar}
\end{figure}

The polar scans are originally $400\times3768$.
For the methods based on polar scans (\ref{V1}, \ref{V3}, \ref{V4}) we suppress the first $60$ range bins by setting them to zero.
This is motivated in~\cref{fig:polar}, where the early returns are likely dominated by the ego-vehicle itself and/or ground returns.
In practice, we found that this boosted performance for all methods.
For quicker processing, we then rescale the $400\times3768$ polar to $400\times512$ -- scaling only range.

\subsubsection{First baseline (\ref{V1})}
\label{exp:v1}

For \ref{V1} we reduce the $400\times512$ polar scan to a $1\times512$ vector by average-pooling the azimuth dimension.
This is a longer representation than is used in~\cite{kim2020mulran}, but we will show in~\cref{sec:baseline_opt} that there are no settings for~\ref{V1} which allows it to approach the performance of~\ref{V3} or~\ref{V4}.

\subsubsection{Second baseline (\ref{V2})}
\label{exp:v2}

Here, a Cartesian scan is made $256$ pixels square, and we use a Cartesian pixel resolution of \SI{1.2717}{\metre}.
This means -- with the vehicle in the centre of the Cartesian projection -- that the range is $\frac{256}{2}\times1.2717=\SI{162.7776}{\metre}$.
With a radar range resolution of \SI{4.32}{\centi\metre} (see~\cref{sec:dataset}) for the polar variants above range is given by $3768\times0.0432=\SI{162.7776}{\metre}$.
Thus, \textit{all methods are presented with precisely the same range-bearing information out to the same range}.
In~\cite{jang2023iros}, after the Radon transform, the Sinograms are scaled to \SI{10}{\percent} of their size.
We use \SI{25}{\percent} by default, but explore the effect of this and the Cartesian image size in~\cref{sec:baseline_opt} (where, again, there are no settings which allow \ref{V2} to approach \ref{V3} or \ref{V4}).

\subsubsection{First ablation (\ref{V3})}
\label{exp:v3}

VLAD descriptors are computed with $64$ cluster centres.
Cluster centres themselves are computed from each reference trajectory.
For this we use \texttt{kmeans++}~\cite{arthur2007k} with tolerance \num{1e-4} and running only a single centroid seed.

\subsubsection{Second ablation (\ref{V4})}
\label{exp:v4}
With discrete radial returns in the form $1\times512$ we perform the $512$-point FFT, resulting in frequency response maps of the same shape as the input polar array ($400\times512$).

\begin{figure}
\centering
\includegraphics[width=\columnwidth]{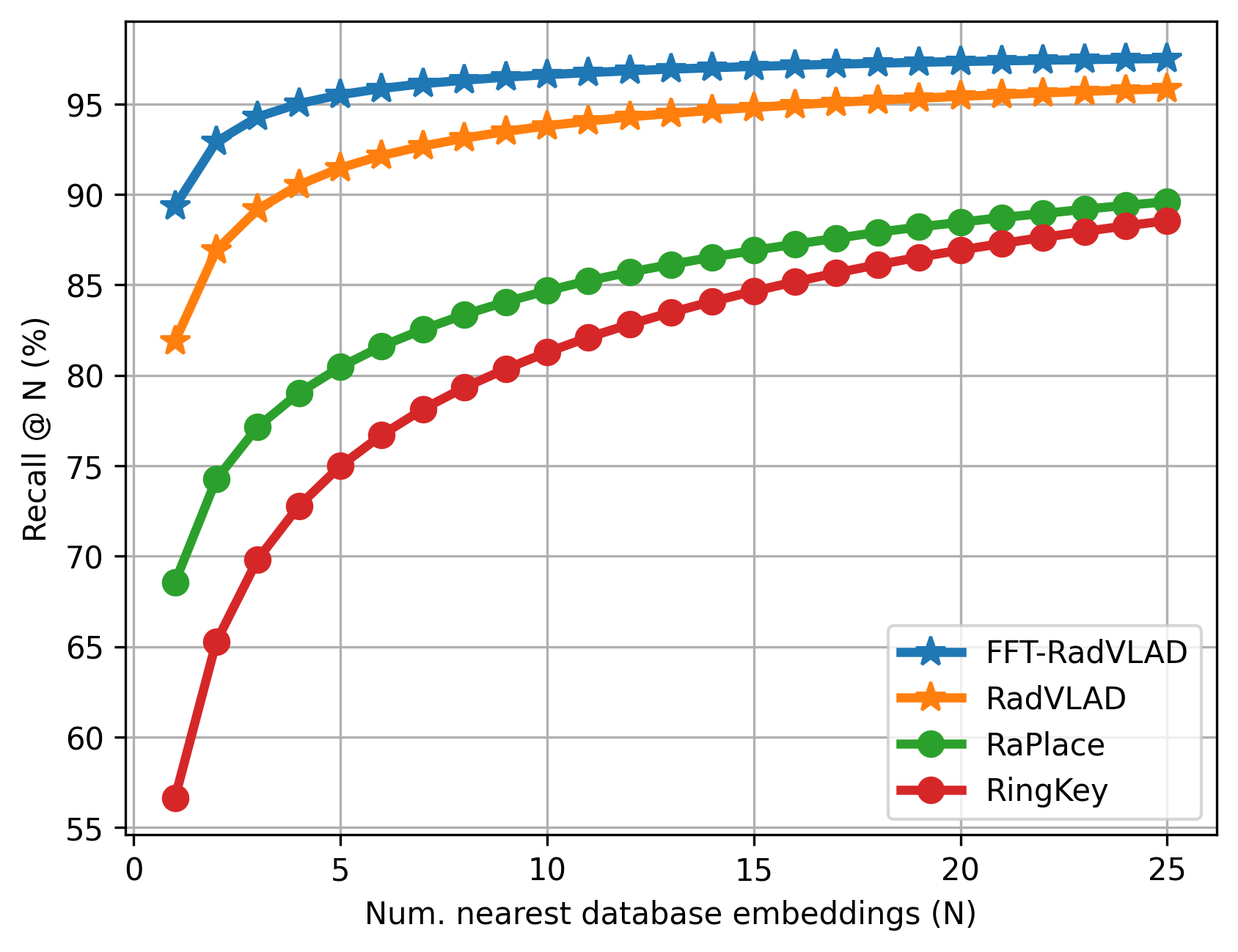}
\caption{
\texttt{Recall@1-50} curves for variants of our method (\ref{V3} and \ref{V4}) and two competitors (\ref{V1} and \ref{V2}).
As a guide to reading this result: for $\mathtt{N}=20$ consider that \textit{Ring Key} has approximately \SI{85}{\percent} \texttt{Recall@20} -- this means that, \SI{85}{\percent} of the time, when a query has $20$ candidate nearest neighbours returned, at least $1$ of them is in fact a nearby place.
Of course, for $\mathtt{N}\rightarrow\infty$ we will eventually return the entire map where we are guaranteed to find a match (\texttt{Recall@}$\infty\rightarrow$\SI{100}{\percent}).
For all \texttt{N}, our \ref{V4} performs best.
}
\label{fig:recall_plot}
\end{figure}

\section{Results}
\label{sec:results}

Experimental results are presented in~\cref{fig:recall_plot,tab:average_results,tab:gridsearch} and are discussed below.
\ifthenelse{\boolean{arxiv}}{Full results are available in~\cref{tab:full_results}.}{}

\begin{table}
\centering
\newcolumntype{g}{>{\columncolor{lightgray}}c}
\resizebox{0.7\columnwidth}{!}{
\begin{tabular}{ccc|g}
\rowcolor{lightgray} \textbf{Azis [px]} & \textbf{Bins [px]} & \textbf{Length [px]} & \texttt{Recall@1} [\%]\\
\hline
50 & 1884 & 128 & 48.31\\
100 & 1884 & 128 & 49.44\\
200 & 1884 & 128 & 49.77\\
400 & 1884 & 128 & 50.23\\
50 & 3768 & 128 & 51.81\\
100 & 3768 & 128 & 52.26\\
200 & 3768 & 128 & 52.37\\
\rowcolor{maroon!30} 400 & 3768 & 128 & 52.93\\
50 & 1884 & 512 & 47.86\\
100 & 1884 & 512 & 48.53\\
200 & 1884 & 512 & 48.87\\
400 & 1884 & 512 & 50.23\\
50 & 3768 & 512 & 51.58\\
100 & 3768 & 512 & 52.26\\
200 & 3768 & 512 & 52.14\\
400 & 3768 & 512 & 52.60\\
\end{tabular}
}
\\[5pt]
\resizebox{0.7\columnwidth}{!}{
\begin{tabular}{ccc|g}
\rowcolor{lightgray} \textbf{Scale [\%]} & \textbf{Res. [m]} & \textbf{Width [px]} & \texttt{Recall@1} [\%]\\
\hline
10 & 1.2717 & 256 & 58.92\\
10 & 0.63585 & 512 & 61.96\\
10 & 2.5424 & 128 & 49.89\\
10 & 0.3178 & 1024 & 64.56\\
20 & 1.2717 & 256 & 62.98\\
20 & 0.63585 & 512 & 64.90\\
20 & 2.5424 & 128 & 58.35\\
20 & 0.3178 & 1024 & 65.69\\
30 & 1.2717 & 256 & 64.56\\
30 & 0.63585 & 512 & 66.25\\
30 & 2.5424 & 128 & 61.63\\
30 & 0.3178 & 1024 & 66.59\\
40 & 1.2717 & 256 & 65.24\\
\rowcolor{maroon!30} 40 & 0.63585 & 512 & \textbf{67.16}\\
40 & 2.5424 & 128 & 63.54\\
40 & 0.3178 & 1024 & 66.93\\
\end{tabular}
}

\caption{
\texttt{Recall@1} (\%) for variations of
\textit{Top}: \ref{V1}.
\textit{Bottom}: \ref{V2} with best-case indicated red.
To explore these combinations of settings we use as localisation and map experiences only \texttt{2019-01-10-11-46-21} and \texttt{2019-01-11-13-24-51}.
}
\label{tab:gridsearch}
\end{table}

\subsection{Establishing an exhaustive open baseline for the \textit{Oxford Radar RobotCar Dataset}}

\cref{tab:average_results} shows the \texttt{Recall@1} as an aggregate for each experience as a query over all possible maps.
\ifthenelse{\boolean{arxiv}}{Full results are available in~\cref{tab:full_results}.}{}
The entries in~\cref{tab:average_results} are read as \ref{V1} / \ref{V2} / \ref{V3} / \ref{V4} or \ding{172} / \ding{173} / \ding{174} / \ding{175} as defined in~\cref{sec:baselines} with \ref{V3} and \ref{V4} or \ding{174} or \ding{175} being the third and fourth entries in each cell, corresponding to variants of our proposed system.
To complement this, \cref{fig:recall_plot} shows the summarised \texttt{Recall@1-50} plots which are averaged (mean) over all of the experience pairs from~\cref{tab:average_results}.
For every one of the $870$ query-reference trajectory pairs that are aggreated in \cref{tab:average_results}, \ref{V4} is the top-performer.
This is reaffirmed in~\cref{fig:recall_plot} which is averaged over all $870$ experiments for the full range of \texttt{N}.

\subsection{Measuring computational efficiency}

Importantly, our \ref{V4} is more computationally efficient than \ref{V2}.
This is shown in~\cref{fig:timing} where time taken to build representations (\ref{V2} representation building includes a polar to Cartesian conversion) is reduced by $\approx$\SI{75}{\percent} and time taken to compute distances between representations by $\approx$\SI{50}{\percent}.

\subsection{Best-case baselines}
\label{sec:baseline_opt}

\cref{tab:gridsearch} \textit{Left} shows an exhaustive variation of three parameters for \ref{V1}.
They are: the maximum radar range (\textbf{Bins [px]}), the number of discrete headings in the polar scan before vector reduction/azimuth pooling (\textbf{Azis [px]}), and the number of discrete range bins in the reduced vector representation (\textbf{Length [px]}).
We are not able to exceed the performance in~\cref{tab:average_results} sufficiently to approach the performance of \ref{V3} or \ref{V4}.

\cref{tab:gridsearch} \textit{Right} similarly shows exhaustive variation of three parameters for \ref{V2}.
They are: the real-world size of a Cartesian pixel (\textbf{Res. [m]}), the number of pixels to a side for the Cartesian scan (\textbf{Width [px]}), and the downscaling factor applied (\textbf{Scale [\%]}) prior to applying the Radon transform.
We are similarly not able to exceed the performance of~\cref{tab:average_results} sufficiently to approach \ref{V3} or \ref{V4}.

\begin{table}
\captionsetup{font=small}
\centering
\newcolumntype{g}{>{\columncolor{lightgray}}c}
\resizebox{\columnwidth}{!}{\begin{tabular}{gcc}
\rowcolor{lightgray} & \texttt{mean(2019-01-*)~~~~} & \texttt{median(2019-01-*)~~}\\
\texttt{2019-01-10-11-46-21} &  56.2 /  68.2 /  81.4 /  \textbf{89.0}  &  57.2 /  69.2 /  83.5 /  \textbf{91.2} \\
\texttt{2019-01-11-13-24-51} &  55.5 /  64.5 /  80.9 /  \textbf{87.0}  &  56.6 /  65.5 /  81.9 /  \textbf{88.1} \\
\texttt{2019-01-14-14-48-55} &  63.7 /  74.1 /  86.7 /  \textbf{93.6}  &  63.9 /  74.3 /  88.0 /  \textbf{94.7} \\
\texttt{2019-01-16-13-09-37} &  58.2 /  67.8 /  83.7 /  \textbf{91.9}  &  59.1 /  69.4 /  85.4 /  \textbf{93.1} \\
\texttt{2019-01-18-12-42-34} &  60.5 /  69.5 /  85.8 /  \textbf{92.0}  &  60.6 /  71.4 /  87.8 /  \textbf{91.6} \\
\texttt{2019-01-10-12-32-52} &  58.1 /  69.7 /  84.8 /  \textbf{91.7}  &  60.0 /  69.6 /  85.6 /  \textbf{93.0} \\
\texttt{2019-01-11-14-02-26} &  61.8 /  73.8 /  86.3 /  \textbf{92.7}  &  62.8 /  74.4 /  87.5 /  \textbf{94.1} \\
\texttt{2019-01-15-12-01-32} &  49.2 /  57.5 /  66.5 /  \textbf{71.5}  &  48.7 /  57.4 /  66.7 /  \textbf{72.4} \\
\texttt{2019-01-16-13-42-28} &  46.6 /  63.5 /  75.4 /  \textbf{85.3}  &  47.7 /  64.1 /  76.0 /  \textbf{85.8} \\
\texttt{2019-01-18-14-14-42} &  60.2 /  74.5 /  87.8 /  \textbf{92.5}  &  61.1 /  75.1 /  88.4 /  \textbf{93.2} \\
\texttt{2019-01-10-14-02-34} &  63.7 /  74.6 /  87.4 /  \textbf{93.0}  &  65.4 /  74.3 /  87.5 /  \textbf{94.0} \\
\texttt{2019-01-11-14-37-14} &  59.8 /  71.0 /  82.5 /  \textbf{89.4}  &  58.2 /  71.2 /  82.9 /  \textbf{90.1} \\
\texttt{2019-01-16-14-15-33} &  45.6 /  61.4 /  74.9 /  \textbf{85.1}  &  46.7 /  61.7 /  75.7 /  \textbf{86.0} \\
\texttt{2019-01-18-14-46-59} &  59.5 /  74.3 /  86.4 /  \textbf{92.5}  &  59.8 /  74.8 /  87.3 /  \textbf{93.1} \\
\texttt{2019-01-14-12-05-52} &  57.6 /  69.6 /  84.3 /  \textbf{91.2}  &  56.9 /  70.5 /  85.5 /  \textbf{92.2} \\
\texttt{2019-01-15-13-06-37} &  59.5 /  69.2 /  85.6 /  \textbf{91.7}  &  60.7 /  70.6 /  87.6 /  \textbf{92.8} \\
\texttt{2019-01-17-11-46-31} &  59.8 /  71.0 /  87.1 /  \textbf{94.0}  &  60.6 /  71.1 /  88.4 /  \textbf{94.5} \\
\texttt{2019-01-18-15-20-12} &  61.2 /  73.0 /  86.6 /  \textbf{92.5}  &  61.5 /  74.0 /  87.2 /  \textbf{93.1} \\
\texttt{2019-01-10-14-50-05} &  59.6 /  72.2 /  83.7 /  \textbf{90.7}  &  60.0 /  72.8 /  83.3 /  \textbf{91.8} \\
\texttt{2019-01-14-12-41-28} &  60.1 /  69.5 /  84.6 /  \textbf{93.0}  &  61.2 /  70.8 /  85.6 /  \textbf{94.6} \\
\texttt{2019-01-15-13-53-14} &  61.9 /  73.2 /  85.1 /  \textbf{91.7}  &  62.9 /  75.5 /  86.5 /  \textbf{92.5} \\
\texttt{2019-01-17-12-48-25} &  56.1 /  68.5 /  84.8 /  \textbf{92.7}  &  56.2 /  69.4 /  86.1 /  \textbf{93.7} \\
\texttt{2019-01-10-15-19-41} &  60.4 /  71.4 /  87.6 /  \textbf{93.8}  &  60.8 /  71.6 /  88.3 /  \textbf{94.2} \\
\texttt{2019-01-14-13-38-21} &  55.8 /  68.6 /  86.1 /  \textbf{92.7}  &  56.1 /  67.7 /  86.7 /  \textbf{94.0} \\
\texttt{2019-01-15-14-24-38} &  60.9 /  72.7 /  85.4 /  \textbf{92.3}  &  61.8 /  74.0 /  86.1 /  \textbf{93.4} \\
\texttt{2019-01-17-13-26-39} &  62.0 /  73.9 /  85.4 /  \textbf{92.4}  &  62.8 /  74.1 /  86.9 /  \textbf{93.1} \\
\texttt{2019-01-11-12-26-55} &  62.2 /  73.5 /  87.3 /  \textbf{93.0}  &  62.7 /  73.1 /  88.6 /  \textbf{94.0} \\
\texttt{2019-01-14-14-15-12} &  61.8 /  71.7 /  84.0 /  \textbf{90.7}  &  64.9 /  73.1 /  86.7 /  \textbf{93.1} \\
\texttt{2019-01-16-11-53-11} &  57.8 /  68.2 /  80.8 /  \textbf{88.6}  &  58.6 /  69.4 /  80.7 /  \textbf{88.3} \\
\texttt{2019-01-17-14-03-00} &  58.8 /  69.2 /  85.6 /  \textbf{94.2}  &  58.6 /  67.1 /  87.0 /  \textbf{95.3} \\
\end{tabular}
}
\caption{
Each entry shows the \texttt{Recall@1} localisation performance (as a percentage, \%) for methods \ding{172}/\ding{173}/\ding{174}/\ding{175} as described in~\cref{sec:baselines}.
Each row corresponds to a trajectory (grey) used as the query experience, and the \texttt{mean} and \texttt{median} columns are then aggregates of this query experience over each of every other trajectory as the reference sequence.
There are $30$ experiences in total, and so each column represents $29$ localisation experiments,  and over $30$ rows we therefore have $29\times30=870$ localisation experiments.
The \textbf{mean over all queries} for \ref{V1} / \ref{V2} / \ref{V3} / \ref{V4} is 56.64 /  68.56 /  81.88 /  \textbf{89.35} and the medians are 57.67 /  69.55 /  83.74 /  \textbf{91.52}.
As can be seen, \ref{V4} (Ours) results in the best performance (\textbf{bold}), in every single case.
}
\label{tab:average_results}
\vspace{-15pt}
\end{table}

\section{Conclusion}

We have presented an open implementation of a simple method for extremely robust radar place recognition.
The learned components of our system (cluster centres) are found in an unsupervised fashion without labels and with only a single hyperparameter (number of cluster centres), meaning that \ref{V3} or \ref{V4} can be extremely easily and quickly included in existing radar localisation pipelines.
Our system was demonstrated over the most exhaustive \textit{Oxford Radar RobotCar Dataset} evaluation to date, and outperforms two other open-sourced competitors.
We trust that the open-sourcing of this system and the full list of results will prompt more research into this problem.

In the future, we will extend the system to other classes of radar, e.g. as in \textit{nuscenes}~\cite{caesar2020nuscenes} as investigated for place recognition in~\cite{cai2022autoplace}.

\section*{Acknowledgements}

This work was supported by EPSRC Programme Grant ``From Sensing to Collaboration'' (EP/V000748/1).
We are also grateful to our partners at Navtech Radar.

\bibliographystyle{IEEEtran}
\bibliography{biblio}

\ifthenelse{\boolean{arxiv}}{
\begin{table*}
\captionsetup{font=small}
\centering
\newcolumntype{g}{>{\columncolor{lightgray}}c}
\resizebox{\textwidth}{!}{\begin{tabular}{g|*{8}{c}}
\rowcolor{lightgray} & \texttt{2019-01-10-11-46-21} & \texttt{2019-01-11-13-24-51} & \texttt{2019-01-14-14-48-55} & \texttt{2019-01-16-13-09-37} & \texttt{2019-01-18-12-42-34} & \texttt{2019-01-10-12-32-52} & \texttt{2019-01-11-14-02-26} & \texttt{2019-01-15-12-01-32}\\
\texttt{2019-01-10-11-46-21} &  -  /  -  /  -  /  -  &  52.6 /  63.9 /  87.0 /  \textbf{93.1}  &  54.1 /  71.4 /  83.1 /  \textbf{93.9}  &  59.1 /  67.5 /  85.9 /  \textbf{94.2}  &  56.2 /  71.1 /  84.4 /  \textbf{90.6}  &  73.6 /  77.1 /  92.0 /  \textbf{96.7}  &  57.0 /  67.6 /  87.2 /  \textbf{94.2}  &  45.6 /  53.3 /  62.3 /  \textbf{71.9} \\
\texttt{2019-01-11-13-24-51} &  54.3 /  63.0 /  82.1 /  \textbf{88.8}  &  -  /  -  /  -  /  -  &  57.1 /  65.3 /  77.8 /  \textbf{86.9}  &  58.7 /  62.5 /  81.6 /  \textbf{87.4}  &  58.6 /  63.7 /  80.5 /  \textbf{87.5}  &  60.2 /  64.8 /  84.4 /  \textbf{89.2}  &  58.2 /  69.5 /  82.1 /  \textbf{89.9}  &  41.3 /  48.9 /  58.6 /  \textbf{68.2} \\
\texttt{2019-01-14-14-48-55} &  63.2 /  74.9 /  86.7 /  \textbf{94.2}  &  61.3 /  69.8 /  84.3 /  \textbf{93.2}  &  -  /  -  /  -  /  -  &  63.0 /  74.4 /  89.3 /  \textbf{94.9}  &  60.4 /  74.2 /  83.6 /  \textbf{92.8}  &  69.2 /  76.6 /  89.4 /  \textbf{92.7}  &  62.1 /  76.5 /  86.6 /  \textbf{91.8}  &  48.6 /  58.6 /  66.0 /  \textbf{70.1} \\
\texttt{2019-01-16-13-09-37} &  55.0 /  68.7 /  84.3 /  \textbf{92.7}  &  59.7 /  65.0 /  82.2 /  \textbf{90.9}  &  57.9 /  69.6 /  84.6 /  \textbf{93.2}  &  -  /  -  /  -  /  -  &  57.7 /  66.2 /  82.3 /  \textbf{92.9}  &  54.9 /  68.8 /  82.7 /  \textbf{93.1}  &  52.9 /  70.3 /  85.4 /  \textbf{92.7}  &  41.5 /  54.7 /  62.7 /  \textbf{70.7} \\
\texttt{2019-01-18-12-42-34} &  55.7 /  69.6 /  84.9 /  \textbf{94.3}  &  55.1 /  59.3 /  85.8 /  \textbf{90.6}  &  52.9 /  64.5 /  81.0 /  \textbf{85.3}  &  65.9 /  70.2 /  87.2 /  \textbf{93.0}  &  -  /  -  /  -  /  -  &  61.9 /  75.0 /  91.6 /  \textbf{95.0}  &  56.7 /  71.9 /  86.8 /  \textbf{93.8}  &  52.0 /  58.5 /  69.8 /  \textbf{73.1} \\
\texttt{2019-01-10-12-32-52} &  63.0 /  76.1 /  87.7 /  \textbf{94.1}  &  53.7 /  62.5 /  86.1 /  \textbf{92.5}  &  57.1 /  71.5 /  81.4 /  \textbf{93.0}  &  56.7 /  71.2 /  84.8 /  \textbf{90.2}  &  63.9 /  69.3 /  88.4 /  \textbf{92.9}  &  -  /  -  /  -  /  -  &  60.0 /  70.4 /  82.2 /  \textbf{92.2}  &  46.8 /  56.7 /  64.2 /  \textbf{69.9} \\
\texttt{2019-01-11-14-02-26} &  55.3 /  71.2 /  84.0 /  \textbf{92.7}  &  59.4 /  66.7 /  84.3 /  \textbf{92.5}  &  58.9 /  75.2 /  87.6 /  \textbf{94.0}  &  58.6 /  71.9 /  85.2 /  \textbf{92.0}  &  63.0 /  72.4 /  83.9 /  \textbf{93.5}  &  60.3 /  74.4 /  87.4 /  \textbf{94.3}  &  -  /  -  /  -  /  -  &  49.0 /  60.6 /  70.0 /  \textbf{74.7} \\
\texttt{2019-01-15-12-01-32} &  49.4 /  57.0 /  64.4 /  \textbf{72.0}  &  44.0 /  50.2 /  62.3 /  \textbf{70.1}  &  50.3 /  58.3 /  68.1 /  \textbf{73.1}  &  47.7 /  60.1 /  65.4 /  \textbf{70.6}  &  50.6 /  59.0 /  68.3 /  \textbf{71.7}  &  46.1 /  60.8 /  66.0 /  \textbf{72.7}  &  49.5 /  60.5 /  67.8 /  \textbf{73.7}  &  -  /  -  /  -  /  - \\
\texttt{2019-01-16-13-42-28} &  47.1 /  63.7 /  76.0 /  \textbf{86.2}  &  42.7 /  56.7 /  73.7 /  \textbf{83.7}  &  42.3 /  60.6 /  73.1 /  \textbf{81.8}  &  50.6 /  67.9 /  78.9 /  \textbf{86.1}  &  49.0 /  63.0 /  77.6 /  \textbf{85.3}  &  48.6 /  60.9 /  80.5 /  \textbf{86.9}  &  41.5 /  59.5 /  75.4 /  \textbf{81.9}  &  36.6 /  52.9 /  60.9 /  \textbf{67.2} \\
\texttt{2019-01-18-14-14-42} &  56.8 /  74.6 /  80.0 /  \textbf{93.4}  &  55.8 /  67.2 /  85.4 /  \textbf{92.8}  &  56.8 /  73.9 /  84.9 /  \textbf{93.9}  &  62.8 /  75.8 /  90.1 /  \textbf{94.9}  &  61.1 /  74.2 /  88.4 /  \textbf{96.1}  &  62.0 /  72.3 /  89.7 /  \textbf{94.2}  &  61.1 /  72.4 /  88.5 /  \textbf{92.7}  &  50.8 /  61.2 /  71.8 /  \textbf{75.7} \\
\texttt{2019-01-10-14-02-34} &  62.9 /  76.2 /  86.1 /  \textbf{94.1}  &  58.0 /  70.5 /  85.5 /  \textbf{91.8}  &  59.1 /  79.3 /  86.3 /  \textbf{94.3}  &  60.7 /  70.1 /  86.8 /  \textbf{90.8}  &  64.7 /  73.5 /  86.4 /  \textbf{93.3}  &  67.6 /  74.1 /  88.1 /  \textbf{95.2}  &  61.4 /  75.3 /  87.9 /  \textbf{93.9}  &  47.3 /  53.0 /  70.1 /  \textbf{73.9} \\
\texttt{2019-01-11-14-37-14} &  55.6 /  72.4 /  83.1 /  \textbf{90.9}  &  63.2 /  69.2 /  80.3 /  \textbf{89.3}  &  56.0 /  73.6 /  84.4 /  \textbf{89.2}  &  58.8 /  69.7 /  84.1 /  \textbf{89.2}  &  60.8 /  71.2 /  82.0 /  \textbf{90.5}  &  59.7 /  72.8 /  81.9 /  \textbf{90.1}  &  63.1 /  75.3 /  84.8 /  \textbf{93.7}  &  46.0 /  56.4 /  62.9 /  \textbf{66.7} \\
\texttt{2019-01-16-14-15-33} &  44.0 /  59.2 /  73.4 /  \textbf{84.4}  &  40.3 /  55.2 /  72.3 /  \textbf{84.7}  &  40.8 /  53.4 /  67.7 /  \textbf{80.5}  &  46.0 /  68.9 /  77.8 /  \textbf{89.0}  &  44.9 /  61.5 /  74.1 /  \textbf{84.7}  &  47.6 /  65.9 /  79.5 /  \textbf{88.5}  &  41.4 /  54.0 /  71.4 /  \textbf{82.6}  &  34.2 /  50.7 /  56.4 /  \textbf{64.7} \\
\texttt{2019-01-18-14-46-59} &  56.9 /  75.7 /  83.6 /  \textbf{94.0}  &  60.7 /  70.9 /  87.3 /  \textbf{93.1}  &  62.5 /  74.7 /  88.5 /  \textbf{94.8}  &  57.4 /  76.0 /  87.3 /  \textbf{92.4}  &  60.7 /  75.5 /  87.5 /  \textbf{94.9}  &  64.9 /  72.2 /  88.9 /  \textbf{95.2}  &  55.2 /  77.8 /  87.6 /  \textbf{92.8}  &  50.1 /  61.2 /  71.4 /  \textbf{75.8} \\
\texttt{2019-01-14-12-05-52} &  57.1 /  69.2 /  82.7 /  \textbf{92.9}  &  61.0 /  68.4 /  84.3 /  \textbf{92.1}  &  56.5 /  70.9 /  83.4 /  \textbf{93.4}  &  57.7 /  72.5 /  81.8 /  \textbf{90.5}  &  54.2 /  66.9 /  80.3 /  \textbf{92.5}  &  57.7 /  67.3 /  85.6 /  \textbf{94.3}  &  59.0 /  71.8 /  83.5 /  \textbf{92.1}  &  48.2 /  59.2 /  65.8 /  \textbf{73.5} \\
\texttt{2019-01-15-13-06-37} &  64.7 /  68.2 /  86.8 /  \textbf{93.1}  &  59.4 /  63.3 /  85.7 /  \textbf{92.9}  &  60.3 /  69.3 /  87.4 /  \textbf{93.8}  &  63.3 /  65.8 /  86.8 /  \textbf{93.1}  &  64.2 /  69.7 /  85.0 /  \textbf{92.6}  &  63.7 /  66.5 /  87.8 /  \textbf{92.5}  &  57.5 /  71.9 /  84.3 /  \textbf{94.2}  &  55.4 /  57.9 /  72.1 /  \textbf{77.4} \\
\texttt{2019-01-17-11-46-31} &  57.4 /  69.7 /  86.5 /  \textbf{91.8}  &  66.1 /  68.3 /  87.8 /  \textbf{94.2}  &  57.6 /  73.1 /  86.8 /  \textbf{94.7}  &  62.0 /  69.3 /  89.3 /  \textbf{94.0}  &  55.2 /  67.9 /  88.6 /  \textbf{94.4}  &  60.0 /  71.8 /  89.3 /  \textbf{94.4}  &  54.3 /  71.0 /  85.0 /  \textbf{94.2}  &  42.4 /  54.4 /  69.2 /  \textbf{74.0} \\
\texttt{2019-01-18-15-20-12} &  56.4 /  70.0 /  85.9 /  \textbf{93.7}  &  54.2 /  64.5 /  81.0 /  \textbf{91.8}  &  63.2 /  77.4 /  86.6 /  \textbf{93.5}  &  57.7 /  68.2 /  87.2 /  \textbf{91.4}  &  59.8 /  69.6 /  87.4 /  \textbf{93.2}  &  63.0 /  76.1 /  91.5 /  \textbf{94.8}  &  49.3 /  72.0 /  80.2 /  \textbf{89.5}  &  47.3 /  53.0 /  71.3 /  \textbf{74.0} \\
\texttt{2019-01-10-14-50-05} &  61.3 /  72.9 /  86.7 /  \textbf{92.8}  &  61.7 /  68.4 /  84.9 /  \textbf{91.6}  &  58.3 /  73.9 /  83.5 /  \textbf{91.2}  &  62.9 /  69.7 /  86.2 /  \textbf{92.2}  &  65.1 /  73.2 /  82.9 /  \textbf{89.3}  &  62.2 /  74.9 /  87.5 /  \textbf{91.7}  &  61.4 /  74.9 /  83.9 /  \textbf{90.7}  &  46.7 /  56.1 /  63.2 /  \textbf{69.4} \\
\texttt{2019-01-14-12-41-28} &  59.9 /  68.8 /  87.5 /  \textbf{94.9}  &  57.8 /  63.5 /  83.9 /  \textbf{92.9}  &  60.3 /  74.2 /  87.6 /  \textbf{94.7}  &  60.6 /  69.0 /  82.1 /  \textbf{93.8}  &  60.6 /  70.2 /  82.2 /  \textbf{93.5}  &  62.3 /  74.0 /  85.2 /  \textbf{94.7}  &  58.9 /  69.4 /  83.4 /  \textbf{92.6}  &  49.5 /  55.6 /  64.8 /  \textbf{73.8} \\
\texttt{2019-01-15-13-53-14} &  60.1 /  71.2 /  86.0 /  \textbf{92.8}  &  61.4 /  69.2 /  86.8 /  \textbf{93.5}  &  61.4 /  74.5 /  85.2 /  \textbf{92.6}  &  61.0 /  70.7 /  85.9 /  \textbf{94.6}  &  63.5 /  72.1 /  84.9 /  \textbf{92.5}  &  63.8 /  75.2 /  86.4 /  \textbf{93.6}  &  60.6 /  72.1 /  85.9 /  \textbf{93.2}  &  50.9 /  56.6 /  69.1 /  \textbf{74.2} \\
\texttt{2019-01-17-12-48-25} &  58.7 /  66.9 /  86.3 /  \textbf{95.2}  &  50.6 /  66.7 /  82.9 /  \textbf{92.4}  &  54.3 /  69.9 /  81.2 /  \textbf{92.8}  &  58.2 /  70.8 /  88.5 /  \textbf{95.1}  &  56.1 /  67.8 /  85.4 /  \textbf{91.3}  &  60.1 /  72.3 /  89.3 /  \textbf{94.7}  &  57.1 /  66.8 /  83.0 /  \textbf{92.5}  &  45.0 /  55.0 /  64.4 /  \textbf{70.5} \\
\texttt{2019-01-10-15-19-41} &  64.0 /  75.1 /  84.5 /  \textbf{93.9}  &  54.6 /  57.2 /  87.9 /  \textbf{95.0}  &  60.4 /  77.1 /  87.3 /  \textbf{93.4}  &  64.4 /  76.3 /  88.2 /  \textbf{93.9}  &  59.3 /  74.6 /  87.2 /  \textbf{93.0}  &  67.5 /  75.6 /  89.0 /  \textbf{94.3}  &  59.3 /  74.0 /  87.3 /  \textbf{94.2}  &  51.5 /  53.5 /  70.9 /  \textbf{76.6} \\
\texttt{2019-01-14-13-38-21} &  51.3 /  67.4 /  81.6 /  \textbf{92.5}  &  55.0 /  58.8 /  81.0 /  \textbf{93.2}  &  63.9 /  69.4 /  90.0 /  \textbf{94.8}  &  51.4 /  73.0 /  85.0 /  \textbf{92.7}  &  47.4 /  64.9 /  74.4 /  \textbf{93.5}  &  57.8 /  67.4 /  86.3 /  \textbf{94.5}  &  53.1 /  63.3 /  84.1 /  \textbf{91.8}  &  40.8 /  51.5 /  65.5 /  \textbf{70.6} \\
\texttt{2019-01-15-14-24-38} &  58.8 /  72.7 /  83.5 /  \textbf{94.5}  &  58.9 /  69.7 /  81.8 /  \textbf{93.1}  &  62.4 /  76.5 /  86.5 /  \textbf{94.8}  &  61.9 /  74.0 /  84.6 /  \textbf{93.6}  &  61.1 /  72.4 /  85.6 /  \textbf{91.5}  &  65.2 /  75.4 /  87.9 /  \textbf{94.7}  &  64.6 /  77.7 /  85.0 /  \textbf{91.4}  &  46.6 /  57.7 /  68.3 /  \textbf{72.6} \\
\texttt{2019-01-17-13-26-39} &  60.7 /  76.9 /  83.0 /  \textbf{91.7}  &  57.8 /  66.0 /  80.8 /  \textbf{91.4}  &  58.5 /  71.7 /  83.5 /  \textbf{93.2}  &  62.6 /  72.7 /  88.6 /  \textbf{94.5}  &  64.0 /  74.1 /  87.3 /  \textbf{91.7}  &  61.5 /  74.9 /  86.2 /  \textbf{95.1}  &  61.0 /  75.6 /  87.1 /  \textbf{94.1}  &  49.0 /  59.4 /  69.3 /  \textbf{75.3} \\
\texttt{2019-01-11-12-26-55} &  63.3 /  75.6 /  88.0 /  \textbf{93.3}  &  62.5 /  66.4 /  87.8 /  \textbf{94.2}  &  61.8 /  76.5 /  88.1 /  \textbf{92.3}  &  65.0 /  74.2 /  88.8 /  \textbf{93.3}  &  64.3 /  75.0 /  84.6 /  \textbf{94.0}  &  66.0 /  75.4 /  89.5 /  \textbf{95.7}  &  63.0 /  79.0 /  88.6 /  \textbf{94.4}  &  43.9 /  57.4 /  66.5 /  \textbf{72.2} \\
\texttt{2019-01-14-14-15-12} &  60.9 /  73.7 /  83.1 /  \textbf{91.1}  &  59.2 /  67.7 /  82.5 /  \textbf{91.1}  &  63.8 /  76.6 /  87.6 /  \textbf{92.9}  &  64.7 /  75.9 /  88.4 /  \textbf{93.8}  &  63.3 /  70.8 /  83.9 /  \textbf{92.8}  &  61.4 /  72.6 /  83.8 /  \textbf{93.2}  &  59.3 /  75.3 /  83.5 /  \textbf{93.3}  &  51.1 /  54.2 /  65.4 /  \textbf{71.4} \\
\texttt{2019-01-16-11-53-11} &  58.0 /  69.4 /  80.3 /  \textbf{88.2}  &  56.6 /  64.6 /  82.4 /  \textbf{86.9}  &  58.3 /  70.0 /  76.9 /  \textbf{88.2}  &  55.2 /  68.9 /  82.6 /  \textbf{88.2}  &  52.0 /  70.2 /  81.1 /  \textbf{86.2}  &  58.8 /  69.4 /  82.3 /  \textbf{88.9}  &  52.8 /  68.0 /  80.4 /  \textbf{86.9}  &  48.8 /  57.8 /  72.1 /  \textbf{75.3} \\
\texttt{2019-01-17-14-03-00} &  62.5 /  73.4 /  88.0 /  \textbf{94.6}  &  58.8 /  70.5 /  86.6 /  \textbf{92.8}  &  57.0 /  68.0 /  89.0 /  \textbf{95.6}  &  58.4 /  71.1 /  89.0 /  \textbf{94.3}  &  59.7 /  72.7 /  85.9 /  \textbf{93.6}  &  58.5 /  69.6 /  84.1 /  \textbf{94.7}  &  54.5 /  67.4 /  83.4 /  \textbf{93.8}  &  47.6 /  57.9 /  70.1 /  \textbf{75.2} \\
\end{tabular}
}
\resizebox{\textwidth}{!}{\begin{tabular}{g|*{8}{c}}
\rowcolor{lightgray} & \texttt{2019-01-16-13-42-28} & \texttt{2019-01-18-14-14-42} & \texttt{2019-01-10-14-02-34} & \texttt{2019-01-11-14-37-14} & \texttt{2019-01-16-14-15-33} & \texttt{2019-01-18-14-46-59} & \texttt{2019-01-14-12-05-52} & \texttt{2019-01-15-13-06-37}\\
\texttt{2019-01-10-11-46-21} &  49.0 /  62.4 /  73.5 /  \textbf{85.3}  &  55.8 /  66.9 /  84.3 /  \textbf{91.6}  &  57.9 /  68.6 /  86.1 /  \textbf{92.8}  &  51.1 /  68.7 /  86.5 /  \textbf{91.3}  &  44.1 /  63.0 /  71.6 /  \textbf{80.9}  &  58.8 /  67.3 /  83.3 /  \textbf{91.4}  &  50.7 /  64.7 /  83.6 /  \textbf{90.5}  &  54.4 /  65.8 /  80.6 /  \textbf{88.0} \\
\texttt{2019-01-11-13-24-51} &  46.5 /  57.8 /  69.8 /  \textbf{80.5}  &  53.9 /  64.5 /  80.2 /  \textbf{86.7}  &  53.8 /  62.2 /  78.9 /  \textbf{86.9}  &  59.5 /  65.0 /  79.0 /  \textbf{86.7}  &  44.5 /  60.6 /  63.9 /  \textbf{78.5}  &  55.6 /  60.6 /  79.1 /  \textbf{85.3}  &  51.7 /  62.4 /  74.5 /  \textbf{84.2}  &  54.5 /  63.0 /  76.9 /  \textbf{83.4} \\
\texttt{2019-01-14-14-48-55} &  50.5 /  65.9 /  74.6 /  \textbf{82.2}  &  60.9 /  75.7 /  85.8 /  \textbf{91.8}  &  55.8 /  74.9 /  81.5 /  \textbf{92.7}  &  64.1 /  74.4 /  85.9 /  \textbf{93.0}  &  49.5 /  66.1 /  67.1 /  \textbf{77.5}  &  63.2 /  69.9 /  85.4 /  \textbf{92.7}  &  62.7 /  70.5 /  82.4 /  \textbf{91.0}  &  58.7 /  72.3 /  84.5 /  \textbf{90.4} \\
\texttt{2019-01-16-13-09-37} &  53.7 /  66.0 /  76.4 /  \textbf{88.3}  &  59.0 /  66.6 /  86.0 /  \textbf{91.5}  &  55.7 /  65.5 /  83.4 /  \textbf{91.5}  &  53.8 /  67.1 /  82.3 /  \textbf{90.4}  &  49.5 /  67.9 /  70.4 /  \textbf{83.6}  &  51.3 /  63.5 /  78.0 /  \textbf{90.0}  &  52.9 /  66.0 /  81.6 /  \textbf{87.5}  &  59.1 /  65.2 /  80.8 /  \textbf{86.7} \\
\texttt{2019-01-18-12-42-34} &  48.0 /  60.2 /  73.9 /  \textbf{80.3}  &  62.0 /  69.1 /  88.8 /  \textbf{93.4}  &  54.0 /  68.5 /  84.0 /  \textbf{90.6}  &  61.2 /  73.4 /  87.4 /  \textbf{91.7}  &  52.9 /  72.6 /  75.0 /  \textbf{84.4}  &  58.4 /  68.1 /  86.1 /  \textbf{93.0}  &  57.5 /  62.3 /  81.1 /  \textbf{89.5}  &  56.9 /  61.7 /  81.7 /  \textbf{88.9} \\
\texttt{2019-01-10-12-32-52} &  46.7 /  61.9 /  70.5 /  \textbf{82.9}  &  59.0 /  71.0 /  82.8 /  \textbf{89.3}  &  63.4 /  67.8 /  85.0 /  \textbf{91.2}  &  59.7 /  67.7 /  82.5 /  \textbf{90.5}  &  48.8 /  60.6 /  68.4 /  \textbf{82.2}  &  59.8 /  66.4 /  83.3 /  \textbf{89.3}  &  55.7 /  66.0 /  81.1 /  \textbf{87.3}  &  53.3 /  65.6 /  78.0 /  \textbf{85.8} \\
\texttt{2019-01-11-14-02-26} &  46.1 /  64.2 /  73.8 /  \textbf{85.4}  &  57.5 /  75.0 /  86.6 /  \textbf{92.3}  &  57.0 /  72.2 /  84.6 /  \textbf{92.7}  &  62.2 /  77.5 /  88.8 /  \textbf{93.1}  &  45.3 /  62.5 /  67.5 /  \textbf{82.0}  &  58.3 /  69.5 /  85.5 /  \textbf{91.3}  &  57.0 /  70.3 /  82.0 /  \textbf{90.1}  &  52.5 /  65.8 /  79.0 /  \textbf{88.4} \\
\texttt{2019-01-15-12-01-32} &  47.7 /  56.4 /  61.5 /  \textbf{66.6}  &  51.1 /  56.5 /  64.8 /  \textbf{71.5}  &  49.7 /  56.0 /  65.4 /  \textbf{70.5}  &  49.7 /  58.4 /  65.7 /  \textbf{69.8}  &  40.8 /  54.5 /  51.2 /  \textbf{61.6}  &  48.2 /  56.2 /  65.4 /  \textbf{72.2}  &  46.1 /  58.6 /  66.3 /  \textbf{71.5}  &  52.1 /  60.1 /  67.9 /  \textbf{72.3} \\
\texttt{2019-01-16-13-42-28} &  -  /  -  /  -  /  -  &  45.3 /  63.7 /  76.9 /  \textbf{84.0}  &  42.3 /  61.1 /  68.4 /  \textbf{81.4}  &  43.6 /  62.3 /  75.0 /  \textbf{85.7}  &  51.1 /  79.7 /  74.4 /  \textbf{89.3}  &  45.9 /  59.2 /  74.6 /  \textbf{83.5}  &  41.4 /  56.2 /  74.0 /  \textbf{81.1}  &  43.1 /  59.0 /  71.7 /  \textbf{79.6} \\
\texttt{2019-01-18-14-14-42} &  54.5 /  65.1 /  78.6 /  \textbf{88.8}  &  -  /  -  /  -  /  -  &  61.1 /  68.1 /  87.2 /  \textbf{93.1}  &  64.7 /  72.7 /  85.4 /  \textbf{91.8}  &  50.9 /  67.2 /  73.9 /  \textbf{85.7}  &  69.6 /  76.5 /  87.4 /  \textbf{92.3}  &  61.4 /  68.0 /  84.1 /  \textbf{89.2}  &  59.1 /  69.1 /  81.8 /  \textbf{90.3} \\
\texttt{2019-01-10-14-02-34} &  54.5 /  68.6 /  75.3 /  \textbf{86.0}  &  62.5 /  68.9 /  87.5 /  \textbf{92.1}  &  -  /  -  /  -  /  -  &  55.1 /  74.3 /  84.6 /  \textbf{91.6}  &  46.2 /  64.6 /  70.5 /  \textbf{80.6}  &  61.5 /  70.4 /  85.1 /  \textbf{89.2}  &  58.3 /  72.0 /  83.7 /  \textbf{88.4}  &  65.6 /  69.2 /  84.9 /  \textbf{90.9} \\
\texttt{2019-01-11-14-37-14} &  48.0 /  63.9 /  72.3 /  \textbf{81.1}  &  62.7 /  71.5 /  81.3 /  \textbf{88.8}  &  55.7 /  65.9 /  79.9 /  \textbf{86.5}  &  -  /  -  /  -  /  -  &  50.3 /  66.1 /  67.9 /  \textbf{82.0}  &  58.4 /  70.8 /  78.3 /  \textbf{89.5}  &  58.9 /  69.1 /  80.9 /  \textbf{85.6}  &  57.3 /  66.8 /  78.4 /  \textbf{85.5} \\
\texttt{2019-01-16-14-15-33} &  54.0 /  79.3 /  75.3 /  \textbf{93.9}  &  42.9 /  55.8 /  69.3 /  \textbf{82.6}  &  42.9 /  57.9 /  72.6 /  \textbf{84.7}  &  42.7 /  63.4 /  75.7 /  \textbf{83.5}  &  -  /  -  /  -  /  -  &  45.5 /  53.6 /  68.9 /  \textbf{78.0}  &  41.5 /  56.0 /  71.7 /  \textbf{82.4}  &  41.1 /  54.2 /  68.2 /  \textbf{79.6} \\
\texttt{2019-01-18-14-46-59} &  49.9 /  64.2 /  75.3 /  \textbf{82.4}  &  65.3 /  76.5 /  87.1 /  \textbf{93.9}  &  59.9 /  75.1 /  84.7 /  \textbf{92.3}  &  63.2 /  76.8 /  83.6 /  \textbf{91.5}  &  48.0 /  69.6 /  72.3 /  \textbf{83.3}  &  -  /  -  /  -  /  -  &  61.1 /  73.2 /  84.9 /  \textbf{91.4}  &  57.7 /  71.0 /  79.4 /  \textbf{89.8} \\
\texttt{2019-01-14-12-05-52} &  49.6 /  60.1 /  72.5 /  \textbf{82.3}  &  59.7 /  66.2 /  83.5 /  \textbf{90.8}  &  59.2 /  69.9 /  82.7 /  \textbf{89.4}  &  56.2 /  69.6 /  81.9 /  \textbf{90.6}  &  51.4 /  60.1 /  68.4 /  \textbf{79.4}  &  54.3 /  67.8 /  78.3 /  \textbf{88.1}  &  -  /  -  /  -  /  -  &  59.2 /  68.8 /  79.7 /  \textbf{91.8} \\
\texttt{2019-01-15-13-06-37} &  49.9 /  62.5 /  76.0 /  \textbf{84.4}  &  61.5 /  69.0 /  85.1 /  \textbf{92.4}  &  60.8 /  72.9 /  87.8 /  \textbf{91.0}  &  57.2 /  68.1 /  84.9 /  \textbf{91.1}  &  45.6 /  66.1 /  70.3 /  \textbf{82.4}  &  59.7 /  66.1 /  81.1 /  \textbf{91.1}  &  63.9 /  72.6 /  89.0 /  \textbf{95.0}  &  -  /  -  /  -  /  - \\
\texttt{2019-01-17-11-46-31} &  52.3 /  63.9 /  76.6 /  \textbf{85.9}  &  52.9 /  71.7 /  85.3 /  \textbf{92.8}  &  56.2 /  72.8 /  87.3 /  \textbf{92.4}  &  62.6 /  70.6 /  87.6 /  \textbf{92.9}  &  50.0 /  69.5 /  74.0 /  \textbf{80.3}  &  59.1 /  67.7 /  80.7 /  \textbf{92.0}  &  60.0 /  64.7 /  83.8 /  \textbf{89.8}  &  51.9 /  60.9 /  81.8 /  \textbf{91.8} \\
\texttt{2019-01-18-15-20-12} &  58.6 /  61.6 /  75.8 /  \textbf{88.2}  &  65.0 /  78.3 /  88.3 /  \textbf{93.0}  &  65.2 /  72.3 /  86.2 /  \textbf{91.9}  &  57.5 /  70.8 /  84.8 /  \textbf{90.8}  &  45.5 /  59.2 /  60.6 /  \textbf{75.5}  &  65.9 /  74.4 /  86.8 /  \textbf{92.8}  &  57.7 /  66.4 /  84.1 /  \textbf{90.2}  &  56.2 /  70.4 /  81.0 /  \textbf{89.4} \\
\texttt{2019-01-10-14-50-05} &  47.2 /  64.3 /  71.3 /  \textbf{77.5}  &  62.0 /  75.4 /  82.9 /  \textbf{91.0}  &  63.6 /  75.5 /  85.9 /  \textbf{92.6}  &  59.7 /  72.9 /  85.8 /  \textbf{90.3}  &  48.1 /  63.0 /  67.7 /  \textbf{79.4}  &  61.4 /  70.0 /  83.0 /  \textbf{88.4}  &  57.0 /  71.2 /  82.8 /  \textbf{89.3}  &  55.4 /  69.0 /  78.7 /  \textbf{85.7} \\
\texttt{2019-01-14-12-41-28} &  53.1 /  64.4 /  76.3 /  \textbf{84.6}  &  54.9 /  66.5 /  80.6 /  \textbf{92.0}  &  57.2 /  70.5 /  83.8 /  \textbf{91.6}  &  58.4 /  70.2 /  84.5 /  \textbf{92.6}  &  52.0 /  65.5 /  68.5 /  \textbf{82.6}  &  56.0 /  64.6 /  80.1 /  \textbf{93.4}  &  56.7 /  64.7 /  81.7 /  \textbf{91.4}  &  58.0 /  64.7 /  82.3 /  \textbf{90.3} \\
\texttt{2019-01-15-13-53-14} &  48.8 /  63.0 /  74.9 /  \textbf{84.3}  &  59.9 /  76.0 /  87.8 /  \textbf{94.2}  &  57.1 /  69.9 /  82.1 /  \textbf{90.8}  &  61.4 /  71.2 /  84.5 /  \textbf{90.4}  &  48.8 /  65.0 /  70.0 /  \textbf{82.2}  &  60.9 /  70.5 /  82.8 /  \textbf{90.2}  &  56.4 /  70.0 /  81.4 /  \textbf{88.3}  &  60.5 /  68.9 /  81.3 /  \textbf{88.9} \\
\texttt{2019-01-17-12-48-25} &  45.4 /  63.0 /  75.7 /  \textbf{83.4}  &  50.4 /  66.9 /  83.1 /  \textbf{91.2}  &  49.6 /  64.9 /  80.7 /  \textbf{87.8}  &  54.1 /  69.3 /  83.5 /  \textbf{90.9}  &  47.7 /  64.6 /  70.0 /  \textbf{81.1}  &  53.7 /  64.1 /  82.0 /  \textbf{90.0}  &  51.1 /  66.3 /  80.7 /  \textbf{87.5}  &  50.9 /  64.3 /  78.8 /  \textbf{86.9} \\
\texttt{2019-01-10-15-19-41} &  54.2 /  62.1 /  77.4 /  \textbf{84.0}  &  66.2 /  73.9 /  92.7 /  \textbf{94.8}  &  63.6 /  70.3 /  86.9 /  \textbf{94.2}  &  55.4 /  75.6 /  84.5 /  \textbf{92.5}  &  43.2 /  64.1 /  71.0 /  \textbf{80.8}  &  60.8 /  72.3 /  87.4 /  \textbf{91.9}  &  52.2 /  69.7 /  83.8 /  \textbf{89.9}  &  64.1 /  70.5 /  85.7 /  \textbf{90.5} \\
\texttt{2019-01-14-13-38-21} &  48.8 /  61.0 /  73.0 /  \textbf{86.4}  &  49.6 /  71.4 /  83.6 /  \textbf{91.6}  &  53.6 /  63.9 /  81.8 /  \textbf{89.6}  &  50.2 /  70.4 /  80.7 /  \textbf{89.1}  &  42.5 /  58.6 /  68.6 /  \textbf{82.2}  &  53.4 /  66.0 /  79.3 /  \textbf{90.4}  &  56.3 /  64.8 /  84.2 /  \textbf{90.3}  &  46.1 /  64.6 /  76.5 /  \textbf{90.9} \\
\texttt{2019-01-15-14-24-38} &  51.3 /  64.6 /  74.0 /  \textbf{86.1}  &  58.0 /  73.8 /  85.6 /  \textbf{92.3}  &  62.7 /  72.9 /  84.8 /  \textbf{90.6}  &  61.9 /  71.5 /  83.9 /  \textbf{89.0}  &  45.6 /  64.1 /  69.0 /  \textbf{84.0}  &  61.4 /  72.3 /  83.4 /  \textbf{91.1}  &  55.3 /  67.1 /  80.3 /  \textbf{86.1}  &  60.8 /  68.7 /  81.8 /  \textbf{88.1} \\
\texttt{2019-01-17-13-26-39} &  50.5 /  67.4 /  74.2 /  \textbf{83.0}  &  59.2 /  73.9 /  86.6 /  \textbf{90.9}  &  58.1 /  74.3 /  81.6 /  \textbf{89.7}  &  59.1 /  72.3 /  83.5 /  \textbf{90.3}  &  49.1 /  68.3 /  70.5 /  \textbf{80.9}  &  58.5 /  69.3 /  80.9 /  \textbf{92.8}  &  54.6 /  67.8 /  78.9 /  \textbf{87.7}  &  56.6 /  69.2 /  80.5 /  \textbf{87.7} \\
\texttt{2019-01-11-12-26-55} &  54.9 /  65.6 /  75.6 /  \textbf{83.4}  &  63.5 /  72.3 /  85.1 /  \textbf{91.5}  &  64.6 /  73.7 /  86.0 /  \textbf{92.5}  &  59.5 /  75.0 /  85.5 /  \textbf{92.5}  &  50.5 /  69.7 /  74.4 /  \textbf{77.6}  &  60.8 /  69.8 /  84.6 /  \textbf{91.2}  &  59.7 /  73.3 /  86.3 /  \textbf{91.9}  &  53.1 /  70.6 /  82.4 /  \textbf{87.6} \\
\texttt{2019-01-14-14-15-12} &  51.2 /  67.2 /  72.6 /  \textbf{87.0}  &  54.8 /  75.3 /  83.5 /  \textbf{93.2}  &  57.9 /  70.3 /  83.8 /  \textbf{89.7}  &  61.4 /  73.7 /  84.0 /  \textbf{93.6}  &  47.9 /  65.6 /  67.2 /  \textbf{83.4}  &  59.2 /  70.0 /  79.4 /  \textbf{92.3}  &  56.6 /  67.8 /  81.7 /  \textbf{90.7}  &  58.6 /  70.1 /  83.4 /  \textbf{89.4} \\
\texttt{2019-01-16-11-53-11} &  48.5 /  64.1 /  72.3 /  \textbf{81.2}  &  56.0 /  63.7 /  80.2 /  \textbf{88.5}  &  56.7 /  63.5 /  81.6 /  \textbf{87.7}  &  60.2 /  68.4 /  80.7 /  \textbf{87.3}  &  46.2 /  65.7 /  63.7 /  \textbf{78.8}  &  60.3 /  67.0 /  79.5 /  \textbf{87.5}  &  59.0 /  63.9 /  77.9 /  \textbf{83.5}  &  50.4 /  61.8 /  76.1 /  \textbf{83.2} \\
\texttt{2019-01-17-14-03-00} &  51.8 /  63.8 /  79.7 /  \textbf{86.8}  &  59.6 /  69.8 /  86.8 /  \textbf{93.7}  &  54.1 /  67.4 /  83.7 /  \textbf{93.2}  &  59.7 /  68.5 /  85.8 /  \textbf{93.5}  &  50.2 /  67.5 /  73.4 /  \textbf{83.8}  &  59.2 /  63.2 /  80.3 /  \textbf{89.7}  &  62.0 /  65.6 /  83.6 /  \textbf{89.3}  &  57.6 /  66.6 /  80.3 /  \textbf{89.4} \\
\end{tabular}
}
\resizebox{\textwidth}{!}{\begin{tabular}{g|*{8}{c}}
\rowcolor{lightgray} & \texttt{2019-01-17-11-46-31} & \texttt{2019-01-18-15-20-12} & \texttt{2019-01-10-14-50-05} & \texttt{2019-01-14-12-41-28} & \texttt{2019-01-15-13-53-14} & \texttt{2019-01-17-12-48-25} & \texttt{2019-01-10-15-19-41} & \texttt{2019-01-14-13-38-21}\\
\texttt{2019-01-10-11-46-21} &  57.9 /  66.9 /  85.0 /  \textbf{92.0}  &  56.9 /  66.7 /  82.3 /  \textbf{92.2}  &  60.0 /  68.5 /  85.4 /  \textbf{93.2}  &  49.8 /  68.2 /  84.5 /  \textbf{93.1}  &  54.3 /  67.7 /  85.4 /  \textbf{93.6}  &  62.2 /  67.2 /  87.4 /  \textbf{94.8}  &  56.2 /  69.9 /  84.0 /  \textbf{94.5}  &  52.6 /  68.1 /  86.5 /  \textbf{93.7} \\
\texttt{2019-01-11-13-24-51} &  55.7 /  65.0 /  79.2 /  \textbf{86.8}  &  54.8 /  66.9 /  76.9 /  \textbf{83.9}  &  55.0 /  64.4 /  80.0 /  \textbf{85.2}  &  52.9 /  60.2 /  79.4 /  \textbf{87.4}  &  56.4 /  65.6 /  81.6 /  \textbf{88.3}  &  58.7 /  64.2 /  80.7 /  \textbf{88.0}  &  57.1 /  65.8 /  81.0 /  \textbf{86.3}  &  57.5 /  62.7 /  80.3 /  \textbf{89.4} \\
\texttt{2019-01-14-14-48-55} &  58.7 /  72.8 /  84.9 /  \textbf{90.7}  &  59.7 /  71.7 /  86.1 /  \textbf{90.0}  &  57.7 /  71.2 /  80.9 /  \textbf{91.7}  &  57.0 /  71.6 /  86.4 /  \textbf{92.8}  &  61.1 /  74.4 /  88.6 /  \textbf{95.5}  &  66.4 /  76.9 /  86.6 /  \textbf{94.8}  &  60.1 /  77.0 /  86.6 /  \textbf{95.0}  &  70.6 /  78.4 /  91.9 /  \textbf{95.3} \\
\texttt{2019-01-16-13-09-37} &  54.9 /  63.6 /  80.4 /  \textbf{88.9}  &  57.7 /  63.6 /  81.5 /  \textbf{90.7}  &  54.9 /  66.6 /  81.1 /  \textbf{88.7}  &  53.5 /  65.9 /  84.4 /  \textbf{92.5}  &  54.3 /  70.3 /  83.2 /  \textbf{92.0}  &  57.0 /  69.6 /  85.0 /  \textbf{93.2}  &  56.1 /  69.2 /  81.6 /  \textbf{90.3}  &  58.1 /  69.6 /  84.4 /  \textbf{93.5} \\
\texttt{2019-01-18-12-42-34} &  53.0 /  70.3 /  83.5 /  \textbf{88.9}  &  56.2 /  64.2 /  79.6 /  \textbf{86.4}  &  57.0 /  70.8 /  86.0 /  \textbf{91.1}  &  54.0 /  70.6 /  83.5 /  \textbf{91.6}  &  61.9 /  71.1 /  85.5 /  \textbf{92.1}  &  62.5 /  73.4 /  90.0 /  \textbf{93.8}  &  60.0 /  73.0 /  84.2 /  \textbf{91.7}  &  62.0 /  69.1 /  85.0 /  \textbf{92.8} \\
\texttt{2019-01-10-12-32-52} &  55.2 /  68.8 /  84.4 /  \textbf{88.4}  &  55.3 /  67.8 /  83.4 /  \textbf{91.3}  &  56.6 /  69.5 /  79.0 /  \textbf{89.9}  &  53.2 /  70.3 /  83.2 /  \textbf{91.1}  &  54.6 /  73.1 /  88.0 /  \textbf{91.5}  &  63.4 /  70.6 /  87.3 /  \textbf{90.8}  &  53.8 /  71.2 /  86.3 /  \textbf{92.1}  &  61.3 /  68.9 /  87.5 /  \textbf{92.2} \\
\texttt{2019-01-11-14-02-26} &  55.1 /  66.8 /  84.4 /  \textbf{91.6}  &  61.3 /  70.8 /  84.3 /  \textbf{89.7}  &  58.1 /  71.0 /  81.4 /  \textbf{89.5}  &  55.7 /  69.6 /  83.9 /  \textbf{91.6}  &  58.2 /  73.6 /  81.4 /  \textbf{89.6}  &  64.4 /  71.2 /  85.5 /  \textbf{91.2}  &  57.4 /  72.8 /  84.2 /  \textbf{92.0}  &  63.9 /  76.2 /  86.0 /  \textbf{94.3} \\
\texttt{2019-01-15-12-01-32} &  42.8 /  52.5 /  66.0 /  \textbf{71.4}  &  49.1 /  56.3 /  65.5 /  \textbf{70.5}  &  46.6 /  56.4 /  65.7 /  \textbf{72.0}  &  46.2 /  57.2 /  64.9 /  \textbf{72.1}  &  47.3 /  59.8 /  69.5 /  \textbf{72.2}  &  47.5 /  58.6 /  65.8 /  \textbf{72.0}  &  50.2 /  56.7 /  64.9 /  \textbf{71.5}  &  53.7 /  60.7 /  68.8 /  \textbf{71.4} \\
\texttt{2019-01-16-13-42-28} &  41.9 /  59.9 /  73.5 /  \textbf{81.7}  &  47.2 /  59.6 /  72.6 /  \textbf{83.2}  &  41.5 /  60.7 /  75.4 /  \textbf{81.8}  &  46.8 /  63.3 /  73.1 /  \textbf{83.5}  &  44.0 /  61.7 /  75.4 /  \textbf{84.7}  &  50.1 /  60.0 /  77.3 /  \textbf{85.8}  &  41.4 /  62.1 /  74.3 /  \textbf{81.8}  &  50.5 /  60.0 /  75.9 /  \textbf{85.7} \\
\texttt{2019-01-18-14-14-42} &  55.7 /  74.5 /  86.1 /  \textbf{92.8}  &  65.5 /  77.6 /  88.0 /  \textbf{94.7}  &  62.6 /  73.2 /  83.8 /  \textbf{92.6}  &  55.4 /  72.3 /  86.1 /  \textbf{93.1}  &  56.1 /  73.1 /  89.2 /  \textbf{94.3}  &  59.5 /  71.8 /  85.9 /  \textbf{93.9}  &  62.4 /  75.8 /  88.9 /  \textbf{92.7}  &  57.7 /  73.2 /  88.6 /  \textbf{95.0} \\
\texttt{2019-01-10-14-02-34} &  58.0 /  68.6 /  85.9 /  \textbf{92.6}  &  61.9 /  70.0 /  87.0 /  \textbf{90.0}  &  66.8 /  74.7 /  86.4 /  \textbf{93.6}  &  59.1 /  72.1 /  86.1 /  \textbf{92.2}  &  57.8 /  74.3 /  86.3 /  \textbf{92.8}  &  68.0 /  72.8 /  88.5 /  \textbf{93.7}  &  66.0 /  75.4 /  88.4 /  \textbf{93.8}  &  63.8 /  75.9 /  90.4 /  \textbf{95.2} \\
\texttt{2019-01-11-14-37-14} &  60.9 /  68.3 /  85.2 /  \textbf{89.6}  &  56.0 /  70.7 /  81.5 /  \textbf{86.1}  &  54.8 /  68.8 /  81.3 /  \textbf{87.5}  &  54.8 /  65.9 /  79.1 /  \textbf{89.3}  &  59.9 /  71.6 /  81.9 /  \textbf{90.0}  &  59.2 /  73.2 /  83.6 /  \textbf{90.7}  &  57.3 /  71.2 /  82.0 /  \textbf{90.0}  &  60.7 /  76.3 /  84.8 /  \textbf{90.0} \\
\texttt{2019-01-16-14-15-33} &  44.8 /  62.6 /  73.4 /  \textbf{82.6}  &  41.2 /  51.8 /  68.5 /  \textbf{79.3}  &  41.4 /  56.8 /  73.1 /  \textbf{83.3}  &  40.9 /  58.2 /  69.9 /  \textbf{82.9}  &  41.5 /  62.8 /  71.9 /  \textbf{87.1}  &  44.6 /  61.9 /  77.7 /  \textbf{88.4}  &  43.3 /  54.6 /  74.4 /  \textbf{78.4}  &  45.1 /  57.4 /  72.9 /  \textbf{82.7} \\
\texttt{2019-01-18-14-46-59} &  53.7 /  73.5 /  83.7 /  \textbf{92.6}  &  64.0 /  75.1 /  88.0 /  \textbf{95.2}  &  58.2 /  70.8 /  86.7 /  \textbf{92.2}  &  56.3 /  69.1 /  85.1 /  \textbf{92.2}  &  60.4 /  71.9 /  88.1 /  \textbf{93.2}  &  62.4 /  74.2 /  87.3 /  \textbf{92.6}  &  53.4 /  73.8 /  81.2 /  \textbf{91.3}  &  58.9 /  73.8 /  86.4 /  \textbf{92.3} \\
\texttt{2019-01-14-12-05-52} &  61.3 /  70.4 /  80.8 /  \textbf{90.6}  &  58.1 /  69.1 /  83.2 /  \textbf{90.3}  &  58.3 /  69.4 /  84.4 /  \textbf{91.6}  &  59.7 /  67.3 /  83.8 /  \textbf{92.5}  &  60.3 /  70.0 /  87.0 /  \textbf{93.2}  &  56.1 /  70.6 /  86.5 /  \textbf{92.2}  &  50.3 /  70.0 /  81.9 /  \textbf{88.3}  &  62.9 /  69.9 /  85.8 /  \textbf{93.4} \\
\texttt{2019-01-15-13-06-37} &  57.5 /  69.0 /  85.3 /  \textbf{91.9}  &  59.7 /  69.6 /  83.9 /  \textbf{90.7}  &  61.5 /  66.2 /  85.8 /  \textbf{92.1}  &  53.1 /  66.4 /  83.8 /  \textbf{89.0}  &  63.3 /  68.6 /  88.5 /  \textbf{94.3}  &  61.5 /  71.8 /  87.8 /  \textbf{92.9}  &  61.7 /  75.3 /  85.6 /  \textbf{92.2}  &  65.6 /  72.5 /  86.5 /  \textbf{94.6} \\
\texttt{2019-01-17-11-46-31} &  -  /  -  /  -  /  -  &  52.8 /  68.0 /  86.7 /  \textbf{92.6}  &  55.6 /  65.2 /  86.3 /  \textbf{93.0}  &  54.1 /  69.7 /  86.3 /  \textbf{93.5}  &  55.3 /  67.4 /  86.8 /  \textbf{93.0}  &  58.8 /  72.3 /  88.8 /  \textbf{95.8}  &  47.8 /  68.1 /  82.5 /  \textbf{90.3}  &  58.2 /  66.9 /  88.2 /  \textbf{92.7} \\
\texttt{2019-01-18-15-20-12} &  54.2 /  67.9 /  79.6 /  \textbf{91.7}  &  -  /  -  /  -  /  -  &  54.7 /  64.9 /  82.1 /  \textbf{88.2}  &  47.6 /  66.3 /  75.2 /  \textbf{86.4}  &  56.5 /  69.8 /  80.7 /  \textbf{90.5}  &  57.7 /  63.4 /  84.9 /  \textbf{91.7}  &  63.9 /  71.7 /  86.6 /  \textbf{93.5}  &  61.5 /  72.8 /  86.5 /  \textbf{95.0} \\
\texttt{2019-01-10-14-50-05} &  58.0 /  69.3 /  82.0 /  \textbf{90.4}  &  57.5 /  70.0 /  79.3 /  \textbf{89.0}  &  -  /  -  /  -  /  -  &  54.1 /  64.9 /  83.5 /  \textbf{89.9}  &  64.8 /  75.2 /  88.0 /  \textbf{93.3}  &  57.4 /  73.2 /  81.4 /  \textbf{90.3}  &  62.2 /  76.1 /  82.0 /  \textbf{91.6}  &  64.6 /  72.6 /  85.5 /  \textbf{91.2} \\
\texttt{2019-01-14-12-41-28} &  55.5 /  66.8 /  82.7 /  \textbf{92.6}  &  54.9 /  64.6 /  78.4 /  \textbf{88.7}  &  54.3 /  69.0 /  82.5 /  \textbf{91.6}  &  -  /  -  /  -  /  -  &  62.5 /  75.2 /  88.7 /  \textbf{95.1}  &  58.8 /  66.7 /  83.7 /  \textbf{94.5}  &  56.5 /  68.0 /  83.1 /  \textbf{92.9}  &  57.4 /  70.4 /  85.9 /  \textbf{94.9} \\
\texttt{2019-01-15-13-53-14} &  55.8 /  68.7 /  82.2 /  \textbf{89.5}  &  56.4 /  68.5 /  82.5 /  \textbf{89.5}  &  62.4 /  69.1 /  85.6 /  \textbf{91.8}  &  57.8 /  70.5 /  85.9 /  \textbf{93.2}  &  -  /  -  /  -  /  -  &  57.7 /  69.3 /  82.0 /  \textbf{92.5}  &  56.4 /  75.2 /  86.7 /  \textbf{93.5}  &  65.7 /  73.1 /  86.7 /  \textbf{93.5} \\
\texttt{2019-01-17-12-48-25} &  53.6 /  64.8 /  83.3 /  \textbf{91.9}  &  54.2 /  65.6 /  84.6 /  \textbf{89.9}  &  48.9 /  65.7 /  79.6 /  \textbf{86.7}  &  46.9 /  66.8 /  83.4 /  \textbf{91.0}  &  48.3 /  67.9 /  80.7 /  \textbf{92.2}  &  -  /  -  /  -  /  -  &  55.5 /  68.7 /  84.1 /  \textbf{92.2}  &  58.2 /  68.4 /  85.6 /  \textbf{93.2} \\
\texttt{2019-01-10-15-19-41} &  55.4 /  70.9 /  84.4 /  \textbf{90.2}  &  65.4 /  77.9 /  87.5 /  \textbf{92.6}  &  64.5 /  77.0 /  87.0 /  \textbf{93.0}  &  55.6 /  73.4 /  84.9 /  \textbf{92.3}  &  54.4 /  77.5 /  87.0 /  \textbf{93.5}  &  62.0 /  75.1 /  88.5 /  \textbf{94.6}  &  -  /  -  /  -  /  -  &  61.7 /  75.1 /  86.1 /  \textbf{93.1} \\
\texttt{2019-01-14-13-38-21} &  57.3 /  66.1 /  80.0 /  \textbf{92.1}  &  58.5 /  67.2 /  82.7 /  \textbf{91.9}  &  55.2 /  68.2 /  79.0 /  \textbf{91.8}  &  58.1 /  63.4 /  83.5 /  \textbf{94.0}  &  61.7 /  64.7 /  88.5 /  \textbf{93.7}  &  59.4 /  68.4 /  87.7 /  \textbf{94.0}  &  54.9 /  67.2 /  80.1 /  \textbf{89.8}  &  -  /  -  /  -  /  - \\
\texttt{2019-01-15-14-24-38} &  59.9 /  71.6 /  84.6 /  \textbf{92.8}  &  62.5 /  73.2 /  86.8 /  \textbf{92.8}  &  59.1 /  73.0 /  83.1 /  \textbf{90.8}  &  55.3 /  69.9 /  81.3 /  \textbf{90.3}  &  63.8 /  79.2 /  88.1 /  \textbf{95.5}  &  64.6 /  71.6 /  88.9 /  \textbf{94.0}  &  55.2 /  73.5 /  85.3 /  \textbf{92.0}  &  66.0 /  72.6 /  89.8 /  \textbf{95.0} \\
\texttt{2019-01-17-13-26-39} &  57.0 /  67.9 /  81.7 /  \textbf{89.0}  &  56.5 /  68.2 /  80.8 /  \textbf{89.6}  &  53.7 /  70.1 /  80.5 /  \textbf{90.6}  &  56.7 /  74.8 /  83.9 /  \textbf{92.8}  &  61.2 /  75.6 /  86.2 /  \textbf{91.1}  &  63.0 /  74.7 /  88.4 /  \textbf{95.2}  &  56.0 /  72.6 /  84.5 /  \textbf{91.7}  &  60.6 /  70.5 /  85.2 /  \textbf{93.4} \\
\texttt{2019-01-11-12-26-55} &  66.0 /  75.3 /  89.3 /  \textbf{89.9}  &  61.7 /  69.3 /  85.8 /  \textbf{91.6}  &  64.4 /  71.4 /  85.4 /  \textbf{91.5}  &  55.9 /  72.3 /  84.3 /  \textbf{92.4}  &  59.1 /  72.8 /  85.1 /  \textbf{92.9}  &  67.7 /  76.2 /  87.3 /  \textbf{94.8}  &  55.9 /  72.5 /  84.8 /  \textbf{92.4}  &  63.4 /  70.5 /  88.8 /  \textbf{94.0} \\
\texttt{2019-01-14-14-15-12} &  58.7 /  66.8 /  82.0 /  \textbf{92.4}  &  59.8 /  72.3 /  82.9 /  \textbf{89.3}  &  56.2 /  71.0 /  81.3 /  \textbf{88.8}  &  56.9 /  69.4 /  84.4 /  \textbf{92.1}  &  58.6 /  72.3 /  85.1 /  \textbf{91.2}  &  61.3 /  73.1 /  86.9 /  \textbf{94.0}  &  57.3 /  73.9 /  87.1 /  \textbf{94.1}  &  71.9 /  80.3 /  90.5 /  \textbf{96.7} \\
\texttt{2019-01-16-11-53-11} &  57.9 /  67.4 /  78.3 /  \textbf{87.8}  &  51.5 /  68.1 /  78.0 /  \textbf{87.7}  &  51.1 /  64.7 /  80.2 /  \textbf{86.5}  &  55.5 /  69.6 /  79.1 /  \textbf{87.8}  &  57.0 /  70.5 /  82.2 /  \textbf{87.3}  &  57.1 /  65.8 /  82.6 /  \textbf{87.9}  &  50.9 /  70.9 /  81.1 /  \textbf{88.3}  &  59.9 /  66.6 /  79.6 /  \textbf{87.1} \\
\texttt{2019-01-17-14-03-00} &  54.5 /  67.3 /  85.1 /  \textbf{95.1}  &  58.3 /  66.4 /  83.0 /  \textbf{89.4}  &  56.9 /  69.7 /  86.5 /  \textbf{92.8}  &  59.2 /  69.5 /  87.2 /  \textbf{93.5}  &  60.2 /  72.2 /  88.0 /  \textbf{93.3}  &  62.7 /  73.1 /  87.5 /  \textbf{95.1}  &  54.2 /  70.6 /  84.4 /  \textbf{91.5}  &  64.1 /  74.1 /  89.6 /  \textbf{95.3} \\
\end{tabular}
}
\resizebox{\textwidth}{!}{\begin{tabular}{g|*{6}{c}|gg}
\rowcolor{lightgray} & \texttt{2019-01-15-14-24-38} & \texttt{2019-01-17-13-26-39} & \texttt{2019-01-11-12-26-55} & \texttt{2019-01-14-14-15-12} & \texttt{2019-01-16-11-53-11} & \texttt{2019-01-17-14-03-00} & \texttt{mean(2019-01-*)~~~~} & \texttt{median(2019-01-*)~~}\\
\texttt{2019-01-10-11-46-21} &  53.7 /  69.4 /  86.6 /  \textbf{93.7}  &  61.5 /  75.1 /  88.5 /  \textbf{95.7}  &  57.6 /  70.2 /  87.8 /  \textbf{93.8}  &  59.9 /  72.0 /  86.8 /  \textbf{93.7}  &  45.7 /  67.4 /  79.5 /  \textbf{88.0}  &  60.3 /  75.2 /  89.7 /  \textbf{94.9}  &  56.2 /  68.2 /  81.4 /  \textbf{89.0}  &  57.2 /  69.2 /  83.5 /  \textbf{91.2} \\
\texttt{2019-01-11-13-24-51} &  54.5 /  64.2 /  80.8 /  \textbf{88.8}  &  54.0 /  64.3 /  83.2 /  \textbf{87.3}  &  58.6 /  66.7 /  79.0 /  \textbf{87.6}  &  59.0 /  67.2 /  82.9 /  \textbf{89.1}  &  46.1 /  55.6 /  74.0 /  \textbf{80.6}  &  60.8 /  69.1 /  85.3 /  \textbf{88.6}  &  55.5 /  64.5 /  80.9 /  \textbf{87.0}  &  56.6 /  65.5 /  81.9 /  \textbf{88.1} \\
\texttt{2019-01-14-14-48-55} &  63.8 /  75.4 /  88.2 /  \textbf{95.3}  &  64.1 /  73.1 /  87.9 /  \textbf{94.1}  &  59.2 /  72.1 /  85.0 /  \textbf{92.1}  &  70.3 /  78.4 /  89.3 /  \textbf{95.5}  &  58.1 /  69.8 /  80.3 /  \textbf{88.7}  &  67.0 /  75.6 /  89.5 /  \textbf{95.9}  &  63.7 /  74.1 /  86.7 /  \textbf{93.6}  &  63.9 /  74.3 /  88.0 /  \textbf{94.7} \\
\texttt{2019-01-16-13-09-37} &  58.6 /  70.2 /  85.9 /  \textbf{94.4}  &  61.1 /  69.5 /  84.8 /  \textbf{93.6}  &  57.8 /  67.8 /  80.4 /  \textbf{89.6}  &  62.1 /  69.4 /  88.3 /  \textbf{93.6}  &  50.1 /  60.1 /  76.0 /  \textbf{87.4}  &  59.5 /  70.0 /  86.8 /  \textbf{92.7}  &  58.2 /  67.8 /  83.7 /  \textbf{91.9}  &  59.1 /  69.4 /  85.4 /  \textbf{93.1} \\
\texttt{2019-01-18-12-42-34} &  55.4 /  62.8 /  87.2 /  \textbf{90.6}  &  61.3 /  74.2 /  89.4 /  \textbf{96.3}  &  59.8 /  71.7 /  85.0 /  \textbf{90.2}  &  70.1 /  71.2 /  88.6 /  \textbf{95.4}  &  49.0 /  63.0 /  75.9 /  \textbf{86.6}  &  67.6 /  73.9 /  88.4 /  \textbf{92.7}  &  60.5 /  69.5 /  85.8 /  \textbf{92.0}  &  60.6 /  71.4 /  87.8 /  \textbf{91.6} \\
\texttt{2019-01-10-12-32-52} &  59.7 /  69.6 /  86.3 /  \textbf{92.8}  &  62.4 /  69.6 /  88.9 /  \textbf{94.4}  &  62.0 /  70.7 /  82.9 /  \textbf{90.8}  &  60.4 /  69.0 /  87.5 /  \textbf{93.3}  &  47.3 /  66.6 /  78.3 /  \textbf{85.7}  &  56.8 /  73.1 /  84.9 /  \textbf{93.3}  &  58.1 /  69.7 /  84.8 /  \textbf{91.7}  &  60.0 /  69.6 /  85.6 /  \textbf{93.0} \\
\texttt{2019-01-11-14-02-26} &  63.0 /  75.4 /  87.6 /  \textbf{91.6}  &  64.0 /  73.8 /  87.5 /  \textbf{94.9}  &  66.0 /  75.0 /  87.2 /  \textbf{93.3}  &  62.2 /  76.6 /  87.5 /  \textbf{94.8}  &  53.1 /  68.4 /  78.7 /  \textbf{86.4}  &  62.6 /  73.6 /  89.3 /  \textbf{94.9}  &  61.8 /  73.8 /  86.3 /  \textbf{92.7}  &  62.8 /  74.4 /  87.5 /  \textbf{94.1} \\
\texttt{2019-01-15-12-01-32} &  48.4 /  60.7 /  67.1 /  \textbf{73.0}  &  50.1 /  57.0 /  67.6 /  \textbf{73.5}  &  48.4 /  54.1 /  65.0 /  \textbf{70.3}  &  54.2 /  61.2 /  68.6 /  \textbf{72.9}  &  45.3 /  54.3 /  64.4 /  \textbf{67.4}  &  49.0 /  57.7 /  66.3 /  \textbf{72.0}  &  49.2 /  57.5 /  66.5 /  \textbf{71.5}  &  48.7 /  57.4 /  66.7 /  \textbf{72.4} \\
\texttt{2019-01-16-13-42-28} &  48.2 /  60.8 /  74.6 /  \textbf{84.5}  &  48.9 /  67.0 /  78.6 /  \textbf{89.5}  &  47.2 /  61.6 /  74.3 /  \textbf{83.1}  &  47.1 /  66.6 /  77.7 /  \textbf{87.5}  &  39.4 /  57.7 /  70.0 /  \textbf{79.7}  &  48.9 /  67.2 /  77.3 /  \textbf{87.2}  &  46.6 /  63.5 /  75.4 /  \textbf{85.3}  &  47.7 /  64.1 /  76.0 /  \textbf{85.8} \\
\texttt{2019-01-18-14-14-42} &  59.5 /  75.9 /  89.6 /  \textbf{93.2}  &  59.9 /  74.6 /  88.8 /  \textbf{94.1}  &  63.2 /  71.8 /  86.5 /  \textbf{91.6}  &  63.1 /  75.5 /  88.0 /  \textbf{94.5}  &  53.1 /  70.5 /  82.6 /  \textbf{88.4}  &  62.4 /  78.4 /  91.2 /  \textbf{93.2}  &  60.2 /  74.5 /  87.8 /  \textbf{92.5}  &  61.1 /  75.1 /  88.4 /  \textbf{93.2} \\
\texttt{2019-01-10-14-02-34} &  65.2 /  75.3 /  89.3 /  \textbf{93.8}  &  67.3 /  72.8 /  88.4 /  \textbf{94.2}  &  66.8 /  74.0 /  86.4 /  \textbf{93.3}  &  65.6 /  78.1 /  89.7 /  \textbf{94.8}  &  55.0 /  73.1 /  83.9 /  \textbf{87.6}  &  62.2 /  74.5 /  86.6 /  \textbf{94.5}  &  63.7 /  74.6 /  87.4 /  \textbf{93.0}  &  65.4 /  74.3 /  87.5 /  \textbf{94.0} \\
\texttt{2019-01-11-14-37-14} &  57.6 /  70.0 /  81.3 /  \textbf{88.3}  &  58.8 /  72.4 /  84.3 /  \textbf{92.3}  &  56.9 /  69.7 /  81.9 /  \textbf{89.1}  &  64.8 /  73.9 /  83.9 /  \textbf{91.2}  &  54.9 /  66.4 /  78.4 /  \textbf{83.7}  &  65.9 /  73.3 /  85.2 /  \textbf{92.0}  &  59.8 /  71.0 /  82.5 /  \textbf{89.4}  &  58.2 /  71.2 /  82.9 /  \textbf{90.1} \\
\texttt{2019-01-16-14-15-33} &  44.6 /  58.8 /  72.3 /  \textbf{84.4}  &  49.6 /  59.4 /  80.4 /  \textbf{88.8}  &  48.8 /  61.9 /  76.5 /  \textbf{83.2}  &  42.7 /  63.2 /  74.9 /  \textbf{87.6}  &  37.4 /  61.5 /  67.3 /  \textbf{79.0}  &  50.3 /  63.8 /  77.8 /  \textbf{87.8}  &  45.6 /  61.4 /  74.9 /  \textbf{85.1}  &  46.7 /  61.7 /  75.7 /  \textbf{86.0} \\
\texttt{2019-01-18-14-46-59} &  64.0 /  75.3 /  86.7 /  \textbf{92.6}  &  60.4 /  74.3 /  89.8 /  \textbf{94.9}  &  59.1 /  70.2 /  83.3 /  \textbf{92.0}  &  58.9 /  76.0 /  87.9 /  \textbf{93.7}  &  49.5 /  72.7 /  82.6 /  \textbf{86.9}  &  65.3 /  77.3 /  88.4 /  \textbf{94.6}  &  59.5 /  74.3 /  86.4 /  \textbf{92.5}  &  59.8 /  74.8 /  87.3 /  \textbf{93.1} \\
\texttt{2019-01-14-12-05-52} &  56.8 /  70.5 /  82.1 /  \textbf{90.3}  &  56.0 /  69.7 /  87.4 /  \textbf{93.5}  &  61.6 /  71.7 /  86.2 /  \textbf{92.3}  &  57.0 /  70.5 /  84.8 /  \textbf{93.2}  &  52.5 /  61.9 /  79.1 /  \textbf{85.7}  &  62.1 /  73.0 /  86.1 /  \textbf{92.1}  &  57.6 /  69.6 /  84.3 /  \textbf{91.2}  &  56.9 /  70.5 /  85.5 /  \textbf{92.2} \\
\texttt{2019-01-15-13-06-37} &  64.9 /  69.9 /  88.3 /  \textbf{94.2}  &  59.9 /  71.8 /  87.4 /  \textbf{93.6}  &  60.6 /  71.4 /  82.9 /  \textbf{91.5}  &  60.8 /  74.6 /  90.0 /  \textbf{94.0}  &  49.4 /  60.4 /  76.8 /  \textbf{84.6}  &  61.7 /  66.9 /  87.9 /  \textbf{92.1}  &  59.5 /  69.2 /  85.6 /  \textbf{91.7}  &  60.7 /  70.6 /  87.6 /  \textbf{92.8} \\
\texttt{2019-01-17-11-46-31} &  60.7 /  70.7 /  89.3 /  \textbf{94.6}  &  60.4 /  71.3 /  86.7 /  \textbf{96.4}  &  59.5 /  75.4 /  87.5 /  \textbf{94.4}  &  61.6 /  70.9 /  89.3 /  \textbf{94.6}  &  53.5 /  64.4 /  79.7 /  \textbf{89.3}  &  62.8 /  73.0 /  90.0 /  \textbf{94.5}  &  59.8 /  71.0 /  87.1 /  \textbf{94.0}  &  60.6 /  71.1 /  88.4 /  \textbf{94.5} \\
\texttt{2019-01-18-15-20-12} &  68.8 /  75.0 /  89.9 /  \textbf{93.0}  &  60.7 /  71.6 /  88.6 /  \textbf{93.1}  &  61.1 /  74.0 /  83.8 /  \textbf{90.6}  &  63.3 /  78.4 /  91.5 /  \textbf{95.7}  &  51.3 /  65.2 /  80.1 /  \textbf{89.6}  &  61.8 /  74.0 /  85.7 /  \textbf{93.1}  &  61.2 /  73.0 /  86.6 /  \textbf{92.5}  &  61.5 /  74.0 /  87.2 /  \textbf{93.1} \\
\texttt{2019-01-10-14-50-05} &  59.0 /  73.6 /  84.6 /  \textbf{92.0}  &  61.0 /  69.3 /  82.9 /  \textbf{91.6}  &  59.1 /  71.9 /  83.3 /  \textbf{90.0}  &  60.9 /  74.5 /  83.2 /  \textbf{92.5}  &  54.9 /  69.9 /  79.0 /  \textbf{86.2}  &  62.5 /  74.3 /  89.0 /  \textbf{92.2}  &  59.6 /  72.2 /  83.7 /  \textbf{90.7}  &  60.0 /  72.8 /  83.3 /  \textbf{91.8} \\
\texttt{2019-01-14-12-41-28} &  60.2 /  68.4 /  83.1 /  \textbf{92.1}  &  65.5 /  73.4 /  86.7 /  \textbf{97.1}  &  61.9 /  70.1 /  84.5 /  \textbf{93.8}  &  64.8 /  73.3 /  88.4 /  \textbf{95.4}  &  47.8 /  60.6 /  76.3 /  \textbf{83.9}  &  60.5 /  71.5 /  88.4 /  \textbf{95.8}  &  60.1 /  69.5 /  84.6 /  \textbf{93.0}  &  61.2 /  70.8 /  85.6 /  \textbf{94.6} \\
\texttt{2019-01-15-13-53-14} &  60.5 /  75.5 /  87.8 /  \textbf{93.3}  &  64.1 /  76.1 /  86.4 /  \textbf{92.8}  &  61.9 /  68.5 /  83.6 /  \textbf{91.4}  &  65.3 /  75.9 /  86.5 /  \textbf{94.6}  &  55.8 /  67.5 /  78.4 /  \textbf{85.6}  &  63.9 /  75.6 /  88.1 /  \textbf{92.2}  &  61.9 /  73.2 /  85.1 /  \textbf{91.7}  &  62.9 /  75.5 /  86.5 /  \textbf{92.5} \\
\texttt{2019-01-17-12-48-25} &  53.9 /  67.0 /  86.4 /  \textbf{93.9}  &  59.9 /  71.6 /  88.2 /  \textbf{96.2}  &  54.6 /  67.1 /  85.7 /  \textbf{93.4}  &  57.9 /  71.8 /  84.4 /  \textbf{93.2}  &  47.9 /  60.9 /  75.2 /  \textbf{85.2}  &  62.4 /  72.4 /  88.6 /  \textbf{94.3}  &  56.1 /  68.5 /  84.8 /  \textbf{92.7}  &  56.2 /  69.4 /  86.1 /  \textbf{93.7} \\
\texttt{2019-01-10-15-19-41} &  66.9 /  70.5 /  88.6 /  \textbf{94.3}  &  61.5 /  73.2 /  88.3 /  \textbf{94.8}  &  61.7 /  75.6 /  86.5 /  \textbf{92.8}  &  57.0 /  71.5 /  90.9 /  \textbf{97.0}  &  55.5 /  66.1 /  83.2 /  \textbf{89.7}  &  60.1 /  71.7 /  88.2 /  \textbf{94.0}  &  60.4 /  71.4 /  87.6 /  \textbf{93.8}  &  60.8 /  71.6 /  88.3 /  \textbf{94.2} \\
\texttt{2019-01-14-13-38-21} &  55.3 /  67.3 /  86.5 /  \textbf{93.7}  &  56.9 /  68.1 /  87.0 /  \textbf{94.2}  &  54.9 /  66.8 /  85.9 /  \textbf{94.0}  &  62.0 /  75.1 /  90.0 /  \textbf{95.1}  &  48.7 /  63.6 /  79.5 /  \textbf{84.8}  &  57.3 /  70.6 /  87.9 /  \textbf{94.1}  &  55.8 /  68.6 /  86.1 /  \textbf{92.7}  &  56.1 /  67.7 /  86.7 /  \textbf{94.0} \\
\texttt{2019-01-15-14-24-38} &  -  /  -  /  -  /  -  &  62.1 /  74.0 /  86.8 /  \textbf{93.6}  &  61.3 /  70.2 /  86.1 /  \textbf{93.3}  &  68.5 /  77.3 /  88.6 /  \textbf{94.0}  &  51.1 /  68.0 /  79.8 /  \textbf{87.1}  &  61.8 /  74.0 /  85.6 /  \textbf{93.4}  &  60.9 /  72.7 /  85.4 /  \textbf{92.3}  &  61.8 /  74.0 /  86.1 /  \textbf{93.4} \\
\texttt{2019-01-17-13-26-39} &  59.4 /  71.0 /  85.5 /  \textbf{91.8}  &  -  /  -  /  -  /  -  &  62.8 /  74.1 /  86.9 /  \textbf{93.1}  &  63.4 /  75.3 /  87.2 /  \textbf{95.0}  &  57.4 /  71.4 /  80.2 /  \textbf{87.9}  &  66.8 /  77.5 /  87.5 /  \textbf{94.3}  &  62.0 /  73.9 /  85.4 /  \textbf{92.4}  &  62.8 /  74.1 /  86.9 /  \textbf{93.1} \\
\texttt{2019-01-11-12-26-55} &  65.6 /  75.2 /  88.0 /  \textbf{94.0}  &  67.5 /  75.9 /  89.8 /  \textbf{95.8}  &  -  /  -  /  -  /  -  &  60.9 /  73.1 /  88.6 /  \textbf{93.2}  &  54.1 /  70.6 /  80.0 /  \textbf{87.2}  &  62.7 /  72.9 /  89.9 /  \textbf{94.6}  &  62.2 /  73.5 /  87.3 /  \textbf{93.0}  &  62.7 /  73.1 /  88.6 /  \textbf{94.0} \\
\texttt{2019-01-14-14-15-12} &  65.9 /  75.4 /  86.7 /  \textbf{93.1}  &  65.3 /  73.1 /  87.3 /  \textbf{95.0}  &  59.7 /  70.3 /  83.3 /  \textbf{89.2}  &  -  /  -  /  -  /  -  &  53.2 /  64.7 /  74.1 /  \textbf{81.9}  &  64.9 /  74.8 /  88.4 /  \textbf{94.3}  &  61.8 /  71.7 /  84.0 /  \textbf{90.7}  &  64.9 /  73.1 /  86.7 /  \textbf{93.1} \\
\texttt{2019-01-16-11-53-11} &  57.0 /  64.7 /  79.4 /  \textbf{88.3}  &  58.6 /  72.9 /  81.9 /  \textbf{89.5}  &  51.6 /  63.7 /  78.7 /  \textbf{87.1}  &  59.2 /  70.0 /  80.7 /  \textbf{89.7}  &  -  /  -  /  -  /  -  &  62.5 /  69.4 /  83.2 /  \textbf{88.3}  &  57.8 /  68.2 /  80.8 /  \textbf{88.6}  &  58.6 /  69.4 /  80.7 /  \textbf{88.3} \\
\texttt{2019-01-17-14-03-00} &  55.3 /  66.5 /  87.0 /  \textbf{95.3}  &  62.4 /  75.0 /  89.7 /  \textbf{96.1}  &  58.6 /  67.1 /  85.9 /  \textbf{93.7}  &  62.7 /  71.3 /  87.5 /  \textbf{96.3}  &  54.9 /  66.1 /  77.6 /  \textbf{89.6}  &  -  /  -  /  -  /  -  &  58.8 /  69.2 /  85.6 /  \textbf{94.2}  &  58.6 /  67.1 /  87.0 /  \textbf{95.3} \\
\end{tabular}
}
\caption{
Exhaustive pairwise experience-to-experience place recognition experiments over the \textit{Oxford Radar RobotCar Dataset}.
These are listed in full as the most comprehensively detailed set of results in the area of radar place recognition, and should serve as an important benchmark for future research.
These detailed results are aggregated in~\cref{tab:average_results}.
Each entry shows the \texttt{Recall@1} localisation performance (as a percentage, \%) for methods \ding{172}/\ding{173}/\ding{174}/\ding{175} as described in~\cref{sec:baselines}.
To render this sensibly, this is a four-part table, with the top section representing all $30$ experiences each localised against the first $8$ sequences, the next section for all $30$ experiences localised against the next $9$ sequences, etc.
The last two columns in the fourth section aggregate (as mean/median) performance for one sequence as query against all references.
}
\label{tab:full_results}
\end{table*}
}{}

\end{document}